%% file: DNPS.tex
\documentclass[11pt]{article}
\usepackage[dvips]{graphicx}%
\usepackage{textcomp}
\usepackage{amsbsy}
\usepackage{latexsym}
\usepackage[mathscr]{eucal}
\usepackage{amsfonts,amsmath,amsthm}
\usepackage{amssymb}
\usepackage[usenames]{xcolor}
\usepackage{hyperref}

\usepackage[T5,T1]{fontenc}

\usepackage{mathtools}

\usepackage{fullpage}

\begin{document}

\title
{Weighted variation spaces and approximation by shallow ReLU networks\footnote{This research was supported in part by the NSF grants DMS-2134077 and DMS-2134140 of the NSF MoDL program (RD and RN) as well as the ONR MURI grant N00014-20-1-2787 (RD, RN, and JS). 
RP was supported in part by the NSF Graduate Research Fellowship Program under grant DGE-1747503 while he was with the University of Wisconsin--Madison
and the European Research Council under grant 101020573 while he was with the \'Ecole polytechnique f\'ed\'erale de Lausanne. He is now with the University of California, San Diego. JS was supported in part by the NSF grants DMS-2111387 and CCF-2205004.}}
\author{ 
Ronald DeVore, Robert D. Nowak, Rahul Parhi, and Jonathan W. Siegel}
\hbadness=10000
\vbadness=10000
\newtheorem{lemma}{Lemma}[section]
\newtheorem{prop}[lemma]{Proposition}
\newtheorem{cor}[lemma]{Corollary}
\newtheorem{theorem}[lemma]{Theorem}
\newtheorem{remark}[lemma]{Remark}
\newtheorem{example}[lemma]{Example}
\newtheorem{definition}[lemma]{Definition}
\newtheorem{proper}[lemma]{Properties}
\newtheorem{assumption}[lemma]{Assumption}
\newenvironment{disarray}{\everymath{\displaystyle\everymath{}}\array}{\endarray}

\def\RR{\rm \hbox{I\kern-.2em\hbox{R}}}
\def\NN{\rm \hbox{I\kern-.2em\hbox{N}}}
\def\ZZ{\rm {{\rm Z}\kern-.28em{\rm Z}}}
\def\CC{\rm \hbox{C\kern -.5em {\raise .32ex \hbox{$\scriptscriptstyle
|$}}\kern
-.22em{\raise .6ex \hbox{$\scriptscriptstyle |$}}\kern .4em}}
\def\vp{\varphi}
\def\<{\langle}
\def\>{\rangle}
\def\t{\tilde}
\def\i{\infty}
\def\e{\varepsilon}
\def\sm{\setminus}
\def\nl{\newline}
\def\o{\overline}
\def\wt{\widetilde}
\def\wh{\widehat}
\def\cT{{\cal T}}
\def\cA{{\cal A}}
\def\cI{{\cal I}}
\def\cV{{\cal V}}
\def\cB{{\cal B}}
\def\cF{{\cal F}}
\def\cY{{\cal Y}}

\def\cD{{\cal D}}
\def\cP{{\cal P}}
\def\cJ{{\cal J}}
\def\cM{{\cal M}}
\def\cO{{\cal O}}
\def\Chi{\raise .3ex
\hbox{\large $\chi$}} \def\vp{\varphi}
\def\lsima{\hbox{\kern -.6em\raisebox{-1ex}{$~\stackrel{\textstyle<}{\sim}~$}}\kern -.4em}
\def\lsim{\hbox{\kern -.2em\raisebox{-1ex}{$~\stackrel{\textstyle<}{\sim}~$}}\kern -.2em}
\def\[{\Bigl [}
\def\]{\Bigr ]}
\def\({\Bigl (}
\def\){\Bigr )}
\def\[{\Bigl [}
\def\]{\Bigr ]}
\def\({\Bigl (}
\def\){\Bigr )}
\def\L{\pounds}
\def\pr{{\rm Prob}}
\newcommand{\cs}[1]{{\color{magenta}{#1}}}
\def\ds{\displaystyle}
\def\ev#1{\vec{#1}}     %
\newcommand{\lt}{\ell^{2}(\nabla)}
\def\Supp#1{{\rm supp\,}{#1}}
\def\R{\mathbb{R}}
\def\E{\mathbb{E}}
\def\nl{\newline}
\def\T{{\relax\ifmmode I\!\!\hspace{-1pt}T\else$I\!\!\hspace{-1pt}T$\fi}}
\def\N{\mathbb{N}}
\def\Z{\mathbb{Z}}
\def\N{\mathbb{N}}
\def\Zd{\Z^d}
\def\Q{\mathbb{Q}}
\def\C{\mathbb{C}}
\def\Rd{\R^d}
\def\gsim{\mathrel{\raisebox{-4pt}{$\stackrel{\textstyle>}{\sim}$}}}
\def\sime{\raisebox{0ex}{$~\stackrel{\textstyle\sim}{=}~$}}
\def\lsim{\raisebox{-1ex}{$~\stackrel{\textstyle<}{\sim}~$}}
\def\div{\mbox{ div }}
\def\M{M}  \def\NN{N}                  %
\def\L{{\ell}}               %
\def\Le{{\ell^1}}            %
\def\Lz{{\ell^2}}
\def\Let{{\tilde\ell^1}}     %
\def\Lzt{{\tilde\ell^2}}
\def\Ltw{\ell^\tau^w(\nabla)}
\def\t#1{\tilde{#1}}
\def\la{\lambda}
\def\La{\Lambda}
\def\ga{\gamma}
\def\BV{{\rm BV}}
\def\Ga{\eta}
\def\al{\alpha}
\def\cZ{{\cal Z}}
\def\cA{{\cal A}}
\def\cU{{\cal U}}
\def\argmin{\mathop{\rm argmin}}
\def\argmax{\mathop{\rm argmax}}
\def\prob{\mathop{\rm prob}}

\def\cO{{\cal O}}
\def\cA{{\cal A}}
\def\cC{{\cal C}}
\def\cS{{\cal F}}
\def\bu{{\bf u}}
\def\bz{{\bf z}}
\def\bZ{{\bf Z}}
\def\bI{{\bf I}}
\def\cE{{\cal E}}
\def\cD{{\cal D}}
\def\cG{{\cal G}}
\def\cI{{\cal I}}
\def\cJ{{\cal J}}
\def\cM{{\cal M}}
\def\cN{{\cal N}}
\def\cT{{\cal T}}
\def\cU{{\cal U}}
\def\cV{{\cal V}}
\def\cW{{\cal W}}
\def\cL{{\cal L}}
\def\cB{{\cal B}}
\def\cG{{\cal G}}
\def\cK{{\cal K}}
\def\cX{{\cal X}}
\def\cS{{\cal S}}
\def\cP{{\cal P}}
\def\cQ{{\cal Q}}
\def\cR{{\cal R}}
\def\cU{{\cal U}}
\def\bL{{\bf L}}
\def\bl{{\bf l}}
\def\bK{{\bf K}}
\def\bC{{\bf C}}
\def\X{X\in\{L,R\}}
\def\ph{{\varphi}}
\def\D{{\Delta}}
\def\H{{\cal H}}
\def\bM{{\bf M}}
\def\bx{{\bf x}}
\def\bj{{\bf j}}
\def\bG{{\bf G}}
\def\bP{{\bf P}}
\def\bW{{\bf W}}
\def\bT{{\bf T}}
\def\bV{{\bf V}}
\def\bv{{\bf v}}
\def\bt{{\bf t}}
\def\bz{{\bf z}}
\def\bw{{\bf w}}
\def \span{{\rm span}}
\def \meas {{\rm meas}}
\def\rhom{{\rho^m}}
\def\diff{\hbox{\tiny $\Delta$}}
\def\EE{{\rm Exp}}
\def\lll{\langle}
\def\argmin{\mathop{\rm argmin}}
\def\codim{\mathop{\rm codim}}
\def\rank{\mathop{\rm rank}}

\newcommand{\rob}[1]{{\color{blue}{#1}}}

\def\argmax{\mathop{\rm argmax}}
\def\dJ{\nabla}
\newcommand{\ba}{{\bf a}}
\newcommand{\bb}{{\bf b}}
\newcommand{\bc}{{\bf c}}
\newcommand{\bd}{{\bf d}}
\newcommand{\bs}{{\bf s}}
\newcommand{\bff}{{\bf f}}
\newcommand{\bp}{{\bf p}}
\newcommand{\bg}{{\bf g}}
\newcommand{\by}{{\bf y}}
\newcommand{\br}{{\bf r}}
\newcommand{\be}{\begin{equation}}
\newcommand{\ee}{\end{equation}}
\newcommand{\bea}{$$ \begin{array}{lll}}
\newcommand{\eea}{\end{array} $$}
\def \Vol{\mathop{\rm  Vol}}
\def \mes{\mathop{\rm mes}}
\def \Prob{\mathop{\rm  Prob}}
\def \exp{\mathop{\rm    exp}}
\def \sign{\mathop{\rm   sign}}
\def \sp{\mathop{\rm   span}}
\def \rad{\mathop{\rm   rad}}
\def \vphi{{\varphi}}
\def \csp{\overline \mathop{\rm   span}}
\newcommand{\beqn}{\begin{equation}}
\newcommand{\eeqn}{\end{equation}}
\def\beginproof{\noindent{\bf Proof:}~ }
\def\endproof{\hfill\rule{1.5mm}{1.5mm}\\[2mm]}

\newenvironment{Proof}{\noindent{\bf Proof:}\quad}{\endproof}

\renewcommand{\theequation}{\thesection.\arabic{equation}}
\renewcommand{\thefigure}{\thesection.\arabic{figure}}

\makeatletter
\@addtoreset{equation}{section}
\makeatother

\newcommand\abs[1]{\left|#1\right|}
\newcommand\clos{\mathop{\rm clos}\nolimits}
\newcommand\trunc{\mathop{\rm trunc}\nolimits}
\renewcommand\d{d}
\newcommand\dd{d}
\newcommand\diag{\mathop{\rm diag}}
\newcommand\dist{\mathop{\rm dist}}
\newcommand\diam{\mathop{\rm diam}}
\newcommand\cond{\mathop{\rm cond}\nolimits}
\newcommand\eref[1]{{\rm (\ref{#1})}}
\newcommand{\iref}[1]{{\rm (\ref{#1})}}
\newcommand\Hnorm[1]{\norm{#1}_{H^s([0,1])}}
\def\int{\intop\limits}
\renewcommand\labelenumi{(\roman{enumi})}
\newcommand\lnorm[1]{\norm{#1}_{\ell^2(\Z)}}
\newcommand\Lnorm[1]{\norm{#1}_{L_2([0,1])}}
\newcommand\LR{{L_2(\R)}}
\newcommand\LRnorm[1]{\norm{#1}_\LR}
\newcommand\Matrix[2]{\hphantom{#1}_#2#1}
\newcommand\norm[1]{\left\|#1\right\|}
\newcommand\ogauss[1]{\left\lceil#1\right\rceil}
\newcommand{\QED}{\hfill
\raisebox{-2pt}{\rule{5.6pt}{8pt}\rule{4pt}{0pt}}%
  \smallskip\par}
\newcommand\Rscalar[1]{\scalar{#1}_\R}
\newcommand\scalar[1]{\left(#1\right)}
\newcommand\Scalar[1]{\scalar{#1}_{[0,1]}}
\newcommand\Span{\mathop{\rm span}}
\newcommand\supp{\mathop{\rm supp}}
\newcommand\ugauss[1]{\left\lfloor#1\right\rfloor}
\newcommand\with{\, : \,}
\newcommand\Null{{\bf 0}}
\newcommand\bA{{\bf A}}
\newcommand\bB{{\bf B}}
\newcommand\bR{{\bf R}}
\newcommand\bD{{\bf D}}
\newcommand\bE{{\bf E}}
\newcommand\bF{{\bf F}}
\newcommand\bH{{\bf H}}
\newcommand\bU{{\bf U}}

\newcommand\cH{{\cal H}}
\newcommand\sinc{{\rm sinc}}
\def\enorm#1{| \! | \! | #1 | \! | \! |}

\newcommand{\dm}{\frac{d-1}{d}}

\let\bm\bf
\newcommand{\bbeta}{{\mbox{\boldmath$\beta$}}}
\newcommand{\bal}{{\mbox{\boldmath$\alpha$}}}
\newcommand{\bbi}{{\bm i}}

\def\nnew{\color{Red}}
\def\mnew{\color{Blue}}
\def\wnew{\color{magenta}}

\newcommand{\dI}{\Delta}
\newcommand\aconv{\mathop{\rm absconv}}

\newcommand{\RadonOp}{\cR}

\date{}
\maketitle
  \begin{abstract} 
  We investigate  the  approximation of functions $f$ on a bounded domain $\Omega\subset \R^d$ by the outputs of single-hidden-layer ReLU neural networks of width $n$.  This form of nonlinear $n$-term dictionary approximation has been intensely studied since it is  the simplest case of neural network approximation (NNA).  There are several celebrated approximation results for this form of NNA that introduce novel model classes  of functions on $\Omega$ whose approximation rates do not grow unbounded with the input dimension.  These novel classes include Barron classes, and classes based on sparsity or variation such as the Radon-domain BV classes. The present paper is concerned with the definition of these novel model classes  on   domains $\Omega$.   The current definition of these model  classes    does not depend on  the domain $\Omega$.  A new and more proper definition of   model classes on domains  is given by introducing the concept of  weighted variation spaces.  These new model classes are  intrinsic to the domain itself. The importance of these new model classes is that they are strictly larger than the classical (domain-independent) classes. Yet, it is shown that they maintain the same NNA rates.
   
   \end{abstract}
   \textbf{Keywords---}Neural networks, approximation rates, variation spaces.

  \section{Introduction}
  \label{S:intro}
  Neural networks (NNs) are now the numerical method  of choice for the development of learning algorithms in regression and classification, especially when dealing with functions of $d$ variables with $d$ large.
  It is   therefore  important to understand, through mathematical theory, the reasons for this  success.  In learning,
  we are tasked with approximating an unknown function $f$ on a domain $\Omega\subset \R^d$ from some finite set of data observations of $f$.   Thus, at least part of the success in using NNs for such learning problems, must lie in their ability to effectively  approximate
  the functions of interest. While there is no widespread agreement on
  exactly what are these functions of interest, i.e., which functions are encountered in applications, one can ask to describe exactly which functions are well approximated by NNs. 
  
  In approximation theory, questions of this second type are answered by precisely describing the set of functions
  which have a prescribed  rate of approximation using the proposed method of approximation.  In our case of NN approximation, we could for example ask for a precise characterization of the {\it approximation classes} 
  \be 
  \label{Aclass1}
  \cA^\alpha((\Sigma_n)_{n\ge 0}),X):= \{f:\ \dist(f,\Sigma_n)_X =O(n^{-\alpha}), \ n\ge 1 \},\quad \alpha>0,
  \ee 
  where $X$ is some Banach space of functions in which we measure the error of approximation (e.g. $X=L_p(\Omega)$, $1\le p\le \infty$) and $\Sigma_n$ is the set of functions realized by neural networks with $n$ neurons and a prescribed activation function.
  
  For certain methods of approximation such as polynomial or spline approximation   precise characterizations of the corresponding approximation classes are
  known (see, e.g., \cite{DNonlinear,DLbook}) and are described by a smoothness condition on $f$ or equivalently 
  a statement that a function $f$ is in $\cA^\alpha$ if and only if it can be represented
  as a sum of certain fundamental building blocks called atoms with a specified condition on the coefficients in such a representation.

    In the case of NN approximation the characterization of $\cA^\alpha$ appears to be a difficult  problem, which we are not close to solving.  In order to chip away at the problem of characterizing the approximation classes $\cA^\alpha$, we can, as we do in this paper, look at the simplest case of NN approximation which corresponds to approximation 
  by shallow neural networks (one hidden layer) with ReLU activation.  Even in this simplest case, there are no characterizations of the classes $\cA^\alpha$ save for the case $d=1$
  where the desired characterization is known and given by certain smoothness conditions known as Besov regularity (see, e.g., \cite{DNonlinear}). 
  
  The existing literature on NN approximation,
  briefly described below, gives sufficient conditions to be placed on $f$ which guarantee membership in $\cA^\alpha$. We are not aware of nontrivial necessary conditions being known. The present paper will shed some new light on the quest to characterize the approximation classes $\cA^\alpha$ when $d\ge 2$ by showing certain deficiencies in the present theory of sufficient conditions.  We treat only the case where approximation takes place in $X=L_2(\Omega)$.
  
  We prove two  important new results on shallow NN ReLU approximation in this paper. First, we show that
  the characterization of $\cA^\alpha$ will   depend significantly on the geometry of the domain $\Omega$.  For example, the characterization of $\cA^\alpha$ in the case $\Omega=[0,1]^d$ will be different from that of the case when $\Omega$ is the Euclidean ball $B^d$ of $\R^d$.  This means that domain independent results are   insufficient.
 Secondly, we show that if there is any hope of characterizing $\cA^\alpha$ by requiring that
 $f$ has a certain expansion in terms of the elements from the ReLU dictionary, then this will require   conditions on the coefficients in  such expansions which reflect the distance of the atom in the dictionary to the boundary of $\Omega$.   We believe these two new ingredients will prove to be
 important.  
 
 Our main result gives a sufficient condition for membership in $\cA^\alpha$ that is much weaker than
 those previously known.  These new sufficient conditions take the form of $f$ having a NN dictionary expansion
 with  conditions on the coefficients in such an expansions being weaker for atoms close to the boundary of $\Omega$.   While these new results   still are not proven to characterize the approximation classes $\cA^\alpha$ they give much weaker sufficient conditions to guarantee membership in $\cA^\alpha$.

 We turn now to a brief description of some (but not all) of the recent results on shallow NN approximation using ReLU activation functions.  This brief accounting will serve to frame the new results given in the present paper. 
 
  A large number of papers have been written in recent years that give quantitative bounds on the approximation rates of various model classes of functions when using neural networks.  General accountings of such results can be found in \cite{BGKP, DHP,GK, P}.
  Two types of results have emerged.  The first is to show that  deep NNs with ReLU activation functions (see, e.g.,  \cite{lu2021deep,ZX,JX,Y}) are surprisingly effective in approximating functions from classical model classes such as finite balls in a Sobolev or Besov space when the approximation error is measured in an $L_p(\Omega)$ norm with $1\le p\le \infty$.  While such results are deep and interesting, they do not match the most common  setting of learning in high dimensions ($d$ large) because these model classes necessarily suffer the curse of dimensionality.  Indeed, the approximation rates for such smoothness classes  is
  of the form $O(n^{-s/d})$ with $s$ related to the smoothness assumption on $f$. Here and later $n$ always refers to the number of neurons used in the approximation. Thus, for large values of $d$, membership in such a   model class is  not a realistic assumption to make on the target function $f$ to be learned.  This negativity for classical smoothness as a model class asssumption for $f$ can be ameliorated by assuming that the input variable to $f$ (and hence the data as well) is restricted  by a probability
  measure $\mu$ on $\Omega$ supported on a low-dimensional submanifold.
    
    The second type of approximation  result introduces
  novel high-dimensional model classes for which neural network approximation (NNA) rates do not grow unbounded with the input dimension $d$.  Thus, membership in these new model classes    can   be a realistic model class assumption for learning a function of many variables. The most celebrated examples of such new model classes are the Barron class $\cB^s$, $s>0$,  introduced in \cite{B}.  The set $\cB^s=\cB^s(\R^d)$    consists of all functions $f$ defined on $\R^d$ whose Fourier transform $\hat f$ satisfies
  \be
  \label{barron}
 \|f\|_{\cB^s}:= \int_{\R^d}(1+|\omega|)^s|\hat f(\omega)|\, d\omega < +\infty.
  \ee
  
  The original result of Barron showed that on a bounded domain $\Omega\subset \R^d$, any function $f$ on $\Omega$ which is the restriction of a function from    $\cB^1(\R^d)$ can be approximated in the $L_2(\Omega)$ norm by single-hidden-layer sigmoidal networks with $n$ neurons to an accuracy of $C_\Omega \|f\|_{\cB^1}n^{-1/2}$, $n\ge 1$, where the constant $C_\Omega$ only depends upon the measure of $\Omega$.  Notice that this  approximation rate does not deteriorate with increasing  $d$ in contrast with classical smoothness model classes.  However, one must note that the above definition of Barron spaces depend on $d$ and indeed   get more demanding as $d$ increases.
  
 Barron's result spurred a lot of study and generalizations   over the last decades. In particular, new model classes of functions which have sparse representation of as linear combinations of neural atoms were introduced. In the case of ReLU neurons, the sparsity  class is larger than the (second-order) Barron class and yet preserves the rate of approximation of $n$-term approximation~\cite{ma2022barron}. These spaces based on sparsity are called \emph{variation spaces}~\cite{Bach,kurkova2001bounds,mhaskar2004tractability,SXSharp,SXDict}. We summarize these activities for ReLU neurons in the following two sections.  For the moment, we only wish to focus on the existing theory for these model classes and their approximation rates on domains $\Omega\subset\R^d$. This is the typical setting in applications. The existing theory defines the corresponding model classes on $\R^d$ and then extends the definition to  domains as the restriction of functions defined on $\R^d$.  As such, the theory and corresponding results are in a strong sense \emph{independent of the domain $\Omega$}.
    While this leads to a simple approximation theory on domains, these results never take into consideration 
  the nature of $\Omega$, e.g., its geometry.  
  
  The purpose of the present paper is to show there is a more satisfactory definition of these
    novel model classes on domains $\Omega$ that leads to domain-dependent results  that are stronger than that provided by the existing theory. We call these new model classes \emph{weighted} variation spaces since they generalize the classical variation space for ReLU neurons by introducing a domain-dependent weighting of the ReLU atoms. These new model classes are strictly larger than the existing variation spaces while still maintaining the same rate of approximation of $n$-term approximation. We develop this domain-dependent theory primarily in the case when $\Omega=B^d$ is the Euclidean unit ball in $\R^d$. To indicate how the theory would depend on the domain $\Omega$,   we also consider the domain $Q^d:=[-1,1]^d$ and contrast the difference in this case with that of $B^d$.

While we develop our results only for the case of ReLU neurons, 
  we believe that the techniques developed in this paper can be applied to the case of ReLU$^k$ neurons, $k>1$. We leave the details to future work.  So, for the remainder of this paper the activation function
  is 
  \be 
  \label{RELU}
  \sigma(t)=t_+=\max\{0,t\}.
  \ee 
  
  This paper is organized as follows.  In the next two sections, we review some of the existing results on ReLU neural network approximation.  This will serve to frame the new results proved in this paper.  In \S \ref{S:variation} we introduce our new (domain-dependent) model classes. In \S \ref{S:approxdictionary}, we prove our new approximation results for $\Omega=B^2$ the unit Euclidean ball in $\R^2$.  We separate out this case since it is the simplest setting to understand. The remaining sections of this paper
  formulate and prove our results for $\Omega=B^d$ which is the Euclidean unit ball in $\R^d$. We also contrast how the results change when $\Omega = Q^d$.
  Finally, we discuss the possible significance of these new model classes for the problem of learning from data.
  
 \section{Approximation by shallow ReLU networks}
 \label{S:ReLU}
  In this paper, we concentrate on a very specific case of NNA, namely approximation by single-hidden-layer ReLU NNs, i.e., the activation function $\sigma$
  is given by \eref{RELU}.
     We study neural network approximation  on a given  bounded domain (the closure of an  open connected set) $\Omega$  of $\R^d$. The most natural choices for $\Omega$
  are the unit Euclidean ball $B^d$ of $\R^d$ or  the   $d$-dimensional cube $Q^d:=[-1,1]^d$.  
  The case $\Omega=B^d$  will be the   primary example considered in this paper.  In going further, we let $\|\cdot\|$ denote  the Euclidean norm on $\R^d$.  
  
  We define the ReLU atoms
  \be
  \label{atoms}
  \phi(x;\xi,t):=    \sigma(\xi \cdot x-t)=(\xi\cdot x-t)_+,\quad \xi \in \R^d,\ \|\xi\|=1,\ t\in\R. 
  \ee
 Given the atom $\phi$,  we let 
  \be
  \label{hyperplane}
  H_\phi := \{x\in \Omega: \xi \cdot x=t\}
  \ee
  be its hyperplane cut.
     $H_\phi$  divides $\Omega$ into two regions $H_\phi^\pm$.  The function $\phi$ is identically zero on   the region $H_\phi^-:=\{x\in\Omega: \ \xi\cdot x\le t\}$  and the  linear function $=\xi \cdot x-t$ on the second region $H_\phi^+:=\{x\in\Omega:\ \xi\cdot x>t\}$.  Notice that for some values of $t$, the atom $\phi$ is identically zero on $\Omega$ so that $ H_\phi^- =\Omega$.

   For each $\Omega$, there is a smallest  interval $T=T(\Omega)$ such that for $t\notin T$, the dictionary element $\phi(\cdot;\xi,t)$ is either identically zero on $\Omega$ or a linear function on $\Omega$.  Let $\cD= \cD(\Omega):=\{\phi(\cdot;\xi,t)\}$   be the dictionary of all   atoms $\phi$ for which $t\in T=T(\Omega)$. We are interested in 
  $n$-term approximation from the dictionary $\cD$.  For $n=1,2,\dots$, let $\Sigma_n:=\Sigma_n(\cD)$ be the set of functions of the form
  \be
  \label{nterms}
  S(x)=\sum_{j=1}^n a_j\phi_j(x),\quad x\in \Omega,
  \ee 
  where the $\phi_j$ are chosen arbitrarily from $\cD$ and $a_1,\dots,a_n$ are real numbers.  When $n=0$, we define  $\Sigma_0:=\{0\}$. The functions $S\in\Sigma_n$ are precisely  the functions on $\Omega$  produced by a single-hidden-layer ReLU network with $n$ neurons, i.e., width $n$.  The set $\Sigma_n$ is thus a $(d+2)n$ dimensional parametric nonlinear manifold parameterized by the $\xi_j\in B^d$, $j=1,\dots,n$,    the $t_j\in\R$, $j=1,\dots,n$,  and the coefficients $a_1,\dots,a_n\in\R$.  Note that a given
  $S\in\Sigma_n$  has in general many representations of the form \eref{nterms}.  In other words, the dictionary $\cD(\Omega)$ is redundant.

  The above paragraph tells us that there are two ways to view shallow network approximation with ReLU activation.  One view is that it is a special case of $n$-term approximation from a dictionary of functions.  Another view is that it is a special case of manifold approximation.  Therefore, a proper assessment of this form of NN approximation would be to compare it with other
  approximation methods of either one of these forms.
  
   Approximation by $\Sigma_n$ is one of the
  simplest examples of neural network approximation (NNA).  It is therefore a fundamental problem to completely understand the approximation properties of $\Sigma_n$, $n\ge 1$, i.e., what are the properties of a function $f$ that determine how well $f$ is approximated by the elements of  $\Sigma_n$.  
  In the case $d=1$ and $\Omega$ is an interval, the set $\Sigma_n$ is the space of piecewise linear function with $n$ breakpoints.  In this special case, approximation by $\Sigma_n$ is well understood (see, e.g., \cite{DHP,DNonlinear}).  So, we restrict ourselves to the case $d\ge 2$ in going
  further in this paper.

 For a function $f$ in $L_p(\Omega)$, $1\le p\le\infty$, we define
  \be
  E_n(f)_p:=E_n(f)_{L_p(\Omega)}:=\inf_{S\in\Sigma_n}\|f-S\|_{L_p(\Omega)}.
  \ee
  This is a form of nonlinear approximation since the set $\Sigma_n$ is not a linear space but rather  a nonlinear manifold.   Rightfully, we often put this form of   approximation in competition with other examples of manifold approximation
  (see, e.g., \cite{CDPW,DHP}).

    From the viewpoint of approximation theory, an understanding of the approximation properties of $\Sigma_n$   would seek to precisely  characterize the approximation classes for $\Sigma_n$ approximation.  An approximation class is the collection of all functions whose approximation error decays at a prescribed decay rate.  For example, for a given $\alpha>0$, we seek a characterization of  the set 
  \be
  \label{approxclass}
  \cA^\alpha:= \cA^\alpha((\Sigma_n)_{n\ge 0},L_p(\Omega))
  \ee
   of
  functions $f\in L_p(\Omega)$ for which
  \be
  \label{Aalpha}
  E_n(f)_p
\le M(n+1)^{-\alpha},\quad n=0,1,2,\dots.
\ee
Note that by definition, $\Sigma_0=\{0\}$  and hence $E_0(f)_p=\|f\|_{L_p(\Omega)}$
The smallest value of $M$ for which \eref{Aalpha} holds is defined as $\|f\|_{\cA^\alpha}$. Notice that $\cA^\alpha$ is a quasi-normed linear space. While for most classical methods of linear and nonlinear approximation,
e.g. polynomials, splines, $n$-term wavelets, there is a characterization of the spaces $\cA^\alpha$ (at least for a certain range of $\alpha$), the case for neural network approximation is much different.
There is at present no known
characterization of $\cA^\alpha$ for any value of $\alpha>0$.   There are however many sufficient conditions that guarantee membership in $\cA^\alpha$ (see \cite{DHP}). 
  
Another (less ambitious) viewpoint of approximation by $\Sigma_n$ is to propose model classes $K$, i.e., compact subsets $K\subset L_p(\Omega)$, and study how well the elements of $K$ can be approximated by the elements of $\Sigma_n$.   This leads to the study of
\be
\label{EK}
E_n(K)_p:=\sup_{f\in K} E_n(f)_p,\quad n\ge 0.
\ee
If one comes up with a set $K$ for which $E_n(K)\le Cn^{-\alpha}$, $n\ge 1$, then clearly $K\subset \cA^\alpha$ and we gain some information about  $\cA^\alpha$.
Many interesting approximation results have been proven for various classical model classes $K$ such as Sobolev and Besov balls, however, the best approximation rates are not known in all cases~\cite{DHP}.
These results show no gain in approximation efficiency when compared with more classical  methods of approximation such as those that use splines or wavelets. 
Moreover, these  classical model classes all suffer the curse of dimensionality: smoothness of order $s$ gives rate decay
$E_n(K)_p\ge Cn^{-s/d}$, $n\ge 0$.

One of the celebrated accomplishments in the study of NNA was the introduction of new model classes   $K$ whose NNA rates do not grow unbounded with the input dimension $d$. They also give us information on $\cA^\alpha$.   We discuss these model classes in the next two sections. In going further in this paper
we only treat the case of approximation in $L_2(\Omega)$.  However, the case of $L_p(\Omega)$ approximation has also been well studied (see \cite{SXSharp}).

\section{Novel (non-classical) model classes}
\label{S:nonclassical}
  While the classical model classes based on smoothness all suffer the curse of dimensionality, certain novel model classes $K$
  have been introduced whose rates do not grow unbounded with the input dimension.  The discovery of these novel model classes begin with the celebrated work of Barron \cite{B}.  We have already defined the Barron spaces $\cB^s(\R^d)$  in the introduction.  
    
  Barron's original results on NNA were for sigmoidal activation and the Barron class $\cB^1(\R^d)$ where he showed that functions in this class, when restricted to a domain $\Omega\subset\R^d$, had an $L_2(\Omega)$ approximation rate $n^{-1/2}$, $n\ge 1$.  It was rather straightforward to extend his approach to proving that
  functions in $\cB^2$ had the same approximation rate when using ReLU activation.  Several follow up papers significantly improved on these original results as we now describe.
  
   Notice that the Barron classes are  formulated for functions which are defined on all of $\R^d$.  Given   a bounded domain $\Omega$, it is not obvious how these classes should be defined on $\Omega$. The definition employed in the literature is that 
  the space $\cB^s(\Omega)$ is the set of function $f$ defined on $\Omega$ which are the restriction of a function $F\in \cB^s(\R^d)$ with norm given by
  \be
  \label{Bnorm}
  \|f\|_{\cB^s(\Omega)}:=\inf_{F|_\Omega=f}\|F\|_{\cB^s(\R^d)},\quad s>0.
  \ee

   With this definition, we have
  \be
  \label{firstbarron}
  E_n(U(\cB^2(\Omega))_{L_2(\Omega)}\le C n^{-1/2},\quad n\ge 1,
  \ee
  where $C$ depends only on the diameter and measure of $\Omega$.  
  Here and later we use the notation $U(Y)$ to denote the unit ball of a normed space $Y$.
  This approximation rate was improved over the years starting with Makovoz \cite{Makovoz} and continuing on with the results of \cite{Bach,KB,SXSharp}.  The current best known approximation rate for $n$-term ReLU NNA is
  \be
  \label{currentbarron}
  E_n(U(\cB^2(\Omega))_{L_2(\Omega)}\le C n^{-\frac 1 2 -\frac{3}{2d}},\quad n\ge 1,
  \ee
  where again $C$ depends only on $d$.
We refer the reader to \cite{SXSharp} for a more detailed discussion of these approximation results.  It is still not known if this rate can be improved for the Barron class $\mathcal{B}^2$.

We turn next to a second family of novel model classes for NNA referred to as variation spaces.     Let
$\cD=\cD(\Omega)$ be the dictionary of   ReLU atoms whose hyperplane cut intersects $\Omega$.   Consider  any function $S=\sum_{j=1}^n a_j \phi_j$, i.e., $S\in\Sigma_n$.   Recall that this representation is not unique.  We define

\be
\label{defV}
V(S):= \inf\left\{\sum_{j=1}^n  |a_j|: S=\sum_{j=1}^na_j\phi_j \right\},
\ee
which is  called  the variation of  $S$ with respect to the dictionary $\cD$.

With this notation in hand, we can define a new space $\cV:=\cV(\Omega)=\cV(\Omega,\cD)$ as the set of all $f$ in $L_2(\Omega)$ for which there is a sequence $S_n\in\Sigma_n$, $n\ge 1$,  such that $\|f-S_n\|_{L_2(\Omega)}\to 0$, $n\to\infty$,
and $V(S_n)\le M$, $n\ge 1$.  Throughout the paper, we will use $\cV$ when the domain $\Omega$ and dictionary $\cD$ are clear from the context, and use $\cV(\Omega)$, $\cV(\cD)$, or  $\cV(\Omega,\cD)$ when we want to call attention to the domain and/or dictionary. The smallest $M$ for which this is true is defined as $\|f\|_{\cV(\Omega)}$.   This space is  called
the {\it variation space} of the dictionary $\cD$.  The space $\cV(\Omega)$ is a Banach space with respect to this norm (see \cite{SXDict} for properties of variation spaces).  A fundamental relation between the Barron and variation space is the embedding
\be
\label{containment}    
\|f\|_{\cV(\Omega)}\le C_\Omega \|f\|_{\cB^2(\Omega)},\quad f\in\cB^2(\Omega),
\ee
with $C_\Omega$ the embedding constant  (which depends only on the diameter of $\Omega$).   The space $\cV(\Omega)$ is strictly larger than $\cB^2(\Omega)$. We remark that the variation space $\cV(\Omega)$ has also been introduced under other names such as the $\mathcal{F}_1$ space~\cite{Bach} and the Barron space \cite{ma2022barron}.

 The variation space $\cV(\Omega)$ has been carefully studied and in particular it has been proven that
that (see~\cite{SXSharp})
\be
  \label{vapproxrate}
  E_n(U(\cV(\Omega)))_{L_2(\Omega)}\le C n^{-\frac 1 2 -\frac{3}{2d}},\quad n\ge 1,
  \ee
  where $C$ depends only on $\Omega$ and $d$. This approximation rate also matches the decay rate of the metric entropy of $U(\cV(\Omega))$~\cite{SXSharp}.
  Notice that this gives the bound \eref{currentbarron} and is in fact how approximation rates for the Barron class are proved. The important thing to note here is that $\cV$ is a larger space than $\cB^2$ but the current best known approximation rates (with shallow ReLU NNs) for both of these classes is the same, namely $O( n^{-\frac 1 2 -\frac{3}{2d}})$, $n\ge 1$. 
  
  We remark that the rate \eqref{vapproxrate} has also been obtained in the $L_\infty$-norm (on the sphere) in \cite{Bach} using deep results from geometric discrepancy theory \cite{bourgain1988distribution,bourgain1989approximation,matouvsek1996improved}, although a gap exists in dimensions $d=2,3$, which was apparently overlooked by the author of \cite{Bach}. Recently, this gap has been completely closed and these results have been generalized to ReLU$^k$ networks for all $k \geq 0$ in \cite{siegel2023optimal}. Similar uniform approximation rates have also been obtained using an entirely different method for a smaller class of functions in \cite{mhaskar2020dimension}. Similar results have also been investigated for ReLU networks whose inputs and outputs take values in Banach spaces~\cite{korolev2022two}. In that work, it is shown that the $n$-term approximation rate is bounded by $O(n^{-\frac{1}{2}})$.

 A major breakthrough in the understanding of $\cV(\Omega)$ was made by characterizing membership of a function $f$ in $\cV(\Omega)$ through the smoothness of its Radon transform.  Namely, it was originally proved in \cite{O+} that a function $f$ is in $\cV(\Omega)$ if and only if $f$ has an extension $F$ to all of $\R^d$ such that the Radon transform $\mathcal{R}(F;\xi,t)$  is in a certain
 smoothness space. Properties and generalizations of this notion of smoothness were extensively studied in~\cite{PN,PNDeep,PNMinimax}, giving rise to a new family of Banach spaces, now referred to as the Radon-domain BV spaces. These spaces are denoted by $\mathcal{R}\mathrm{BV}^k$, $k \in \mathbb{N}$.

 The key result of~\cite{PN} is the following \emph{representer theorem} for these spaces. Let $x_i\in\R^d$, $i=1,\dots,m$, and $y_i \in \R$, $i=1,\dots,m$. Then, there always exists a solution to the data-fitting problem
 
 \be
 \label{minprob}
    \min_{f \in \mathcal{R}\mathrm{BV}^k} \sum_{i=1}^m \mathcal{L}(y_i, f(x_i)) + \lambda |f|_{\mathcal{R}\mathrm{BV}^k}
 \ee
 that takes the form of a function $S$ which is the output of a single-hidden-layer neural network with $\leq m$ neurons and ReLU$^{k-1}$ activation functions. Here, $\mathcal{L}$ is any loss function which is lower-semi-continuous in its second argument and $|f|_{\mathcal{R}\mathrm{BV}^k}$ is the semi-norm which defines the $\mathcal{R}\mathrm{BV}^k$ spaces, which measures smoothness in the Radon domain.  The Radon BV spaces are defined on domains $\Omega\subset \R^d$ via restrictions. For the case $k=2$ (which corresponds to shallow ReLU NNs) it has been shown in~\cite[Theorem~6]{PNMinimax} (see also~\cite[Theorem~2~and~Corollary~1]{SXDict}) that
 \be 
 \label{RBVvar}
 \mathcal{R}\mathrm{BV}^2(\Omega)= \cV(\Omega),
 \ee 
 with equivalent norms. It has also been shown that there exists a solution $S$ to \eref{minprob} which  is in $\Sigma_m(\cD)$ on any bounded domain $\Omega\subset \R^d$~\cite[Theorem~5]{PNMinimax}.

\section{Weighted variation model classes}
  \label{S:variation}
 One of the main points of the present paper is that one can derive improved results on approximation by shallow ReLU networks if one considers new model classes that generalize the
 standard variation space by including weights on the atoms.  In this section, we introduce these  new model classes for the case when we want the error of approximation to be taken  in the   $L_2(\Omega)$ norm with  $\Omega$   a bounded domain in $\R^d$.    We begin with the general principle of {\it  weighted variation spaces}.
  
  Let $\cD  $ be the dictionary  of ReLU atoms.  Let $S^{d-1}$ be the boundary of the unit Euclidean ball $B^d$ of $\R^d$.
  That is, $S^{d-1}:=\{\xi\in \R^d: \ \|\xi\|=1\} $.   Any atom $\phi$ in $\cD$ is of the form $\phi(x)=(\xi\cdot x-t)_+$ where   $t\in \R $.  We are interested in the atoms $\phi$ whose hyperplane cut intersects $\Omega$ (since otherwise the atom is identically an affine function).  Accordingly, we define
  \be 
  \label{defZ}
  Z(\Omega):=\{(\xi,t): \xi\in S^{d-1}, \ t\in \R \ {\rm such\  that}\ H_{\phi(\cdot;\xi,t)} \cap \Omega \neq \varnothing\}
  \ee
  and $\bar Z(\Omega)$ its closure in the Euclidean norm.
  Note that whenever $\phi(x)=\phi(x;\xi,t)$, $(\xi, t) \in Z(\Omega)$, is positive for some $x\in \Omega$, it is positive in a neighborhood
  of $x$ and hence $\|\phi\|_{L_2(\Omega)}>0$.
  Given the domain $\Omega$ we define the dictionary
  \be 
  \label{dictionaryOmega}
  \cD(\Omega):= \{\phi(\cdot,\xi,t): \ (\xi,t)\in \bar Z(\Omega)\}.
  \ee 
  The set $\bar Z(\Omega)$ is a compact subset of $S^{d-1} \times \R$.  If we equip $\bar Z(\Omega)$ with the Euclidean norm topology then
  the mapping $(\xi,t)\mapsto \phi(\cdot;\xi,t)$ is a continuous mapping from $\bar Z(\Omega)$ into $L_2(\Omega)$.

  Here is an important observation about the atoms in this dictionary which underlies the improved approximation results of this paper.  While each atom $\phi\in\cD(\Omega)$ is in $L_2(\Omega)$ whenever $\Omega$ is a bounded domain, the   $L_2(\Omega)$ norm of $\phi$ will depend heavily on $\phi$ and $\Omega$.  Namely,
  if the support of $\phi$ lies near the boundary of $\Omega$ then this norm will be small and we
  expect that $\phi$ has a less important role in approximating a given target function $f\in L_2(\Omega)$.   

  As an example, consider the case when $\Omega=B^d$ is the $d$-dimensional Euclidean ball. It is easy to see that the atom $\phi(x)=(\xi\cdot x-t)_+$,  has $L_2(\Omega)$-norm satisfying
  \be  
  \label{normphi}
  \|\phi\|_{L_2(\Omega)}\approx (1-t)^{\frac{3}{2}+ \frac{d-1}{4}}, \quad -1\le t\le 1,
  \ee
  with constants of equivalence depending only on $d$.
Indeed, the $L_\infty(\Omega)$ norm of $\phi$ is $1-t$ and the measure of its support
$\approx (1-t)[\sqrt{1-t}]^{d-1}$.  It follows that the norms of atoms get smaller as $t$ approaches one.

  The compactness of $\bar Z(\Omega)$ implies that the dictionary $\cD(\Omega)$ is a compact subset of $L_2(\Omega)$. Thus, there is another useful characterization of the functions
  in $\cV(\Omega)$. Consider the space $\cM:=\cM(\bar Z(\Omega))$ of all finite (signed) Radon measures on $\bar Z(\Omega)$, equipped with the variation norm $\|\mu\|_{\cM}:=\int _{\bar Z(\Omega)} d |\mu|$.  For $\mu\in\cM$, we introduce the function
  \be
  \label{fmu}
   f_\mu:= \int_{\bar Z(\Omega)} \phi(\cdot;\xi,t)\, d\mu(\xi, t),
  \ee 
  where the integral in \eqref{fmu} can be understood as a Bochner integral (see~\cite[Lemma~3]{SXDict} for more details).
 Then, any $f\in\cV(\Omega)$ has a representation
  \be
  \label{Vrep}
  f= f_\mu,\quad \text{for some }\mu\in\cM.
  \ee 
      This representation is not unique in the sense that different measures $\mu$ can give rise to the same $f$.  It then follows (see \cite{SXDict}) that the $\cV$-norm can be alternatively specified by
  \be 
  \label{Vnorm}
  \|f\|_{\cV}= \inf\{ \|\mu\|_{\cM}: \ f=f_\mu, \ \mu\in\cM\}.
  \ee
  
   In order to simplify the geometry, in going further in this section, we assume that $\Omega$ is a convex subset of $\R^d$  and $\cD:=\cD(\Omega)$. 
  We say that  
  \be 
  \label{weightform}
  w(\xi,t),  \quad (\xi,t)\in \bar Z(\Omega),
  \ee 
  is a weight function if $w$ is a non-negative continuous function on $\bar Z(\Omega)$. Given an atom $\phi(\cdot; \xi, t)$ we will abuse notation and also write $w(\phi)$ or $w(\phi(\cdot; \xi, t))$ for $w(\xi, t)$.

  \vskip .1in
  \noindent

{\bf Admissible Weights:}  {\it Given  a weight function $w$ defined on $\bar Z(\Omega)$, we define 
\be\label{defW}
\tilde \phi(\cdot;\xi,t):= \frac{\phi(\cdot;\xi,t)}{w(\xi,t)}, \quad (\xi,t) \in \bar Z(\Omega),
\ee
where $\tilde \phi(\cdot;\xi,t)$ is defined to be the zero function whenever  $w(\xi,t)=0$.  
We say that the weight function $w$ is admissible for $\Omega$, if the mapping $(\xi,t)\to \tilde \phi(\cdot;\xi,t)$  is continuous as a mapping from $\bar Z(\Omega)$ into $L_2(\Omega)$.  It follows that 
\be 
\label{wa}
\|\tilde \phi(\cdot; \xi, t)\|_{L_2(\Omega)} \leq C_w,
\ee
with $C_w$ an absolute constant.
Notice that if a weight function $w$ is admissible, then any larger weight function $\tilde w$ is also admissible.}
\vskip .1in
 When given an admissible weight $w$, the set of functions 
 \be 
 \label{newdictionary} 
 \cD_w:=\cD_w(\Omega)=\{\tilde\phi(\cdot;\xi,t):\ (\xi,t)\in \bar Z(\Omega)\}.
 \ee    
 is a new dictionary contained  in $L_2(\Omega)$.  Furthermore, this dictionary is compact in $L_2(\Omega)$.  
 We define the weighted variation space $\cV_w:=\cV_w(\Omega)$ to be variation space of this new dictionary $\cD_w$. Since the admissibility conditions ensure that the dictionary $\cD_w$ is compact in $L_2(\Omega)$, we have, from the discussion above, that, for every $f\in \cV_w(\Omega)$, there exists a signed Radon measure $\mu=\mu_f$ on $\bar Z(\Omega)$ such that
\begin{equation}
\label{weightrep}
f= \tilde f_\mu:= \int_{\bar Z(\Omega)} \tilde \phi(\cdot;\xi,t) \,d\mu(\xi,t)\quad {\rm with} \quad  \|f\|_{\cV_w(\Omega)}=\|\tilde f_\mu\|_{\cV_w(\Omega)}=\|\mu_f\|_{\cM}.
\end{equation}
We also clearly have
   \be 
   \label{Vwembed}
   \|f\|_{L_2(\Omega)}\le C_w\|f\|_{\cV_w(\Omega)}, \quad f\in\cV_w(\Omega),
   \ee 
   where $C_w$ is the   constant in \eqref{wa}. We also have that, if $\tilde w\ge w$, then $\cV_{\tilde w}(\Omega)\subset \cV_w(\Omega)$ and $\|f\|_{\cV_{\tilde w}}\le\|f\|_{\cV_w(\Omega)}$ which implies that
   \be
   \label{betterapprox}
   E_n(U(\cV_{\tilde w}(\Omega))) \le E_n(U(\cV_w(\Omega))),\quad n\ge 0,
   \ee
   where we note that $\Sigma_n(\cD) = \Sigma_n(\cD_w)$ for any admissible $w$.

   While $\cV_w(\Omega)$ is defined for any nonnegative weight $w$ which is admissible, there is a particular choice of $w$ which we will consider in this paper. Specifically, we show that the approximation rates derived for shallow ReLU neural networks on the unweighted space $\cV(\Omega)$ actually hold on the larger space $\cV_w(\Omega)$ for a certain collection of  admissible weights $w$. As we will later see, the smallest admissible weight with this property will depend upon the domain $\Omega$. 
   This domain-dependent smallest weight is related to the measure of the intersection of the hyperplane of $\phi$ restricted to $\Omega$.  
   To describe this particular weight $w$  and our new approximation results,  we start with the case $d=2$ where the proofs of approximation rates are simplest to understand.   We consider the two domains  $\Omega= B^2$ and $\Omega=Q^2$. Later, we treat the general cases $\Omega=B^d$, $d\ge 2$.
    We then explain how the same theory carries over to  $\Omega=Q^d$ (see Remark \eref{R:Qd}).

    Variation spaces $\cV(\cD_0)$ are defined as above for any dictionary $\cD_0$ in any Hilbert space
    $H$ provided that the dictionary elements $\psi\in\cD_0$ satisfy $\|\psi\|_H\le \delta$ for a fixed value of $\delta>0$. Given such a dictionary $\cD_0$, we define $\Sigma_n:=\Sigma_n(\cD_0)$  as the set of all functions $S\in H$ that are a linear combination of at most $n$ terms of $\cD_0$.  For any $f\in H$, we define the error of $n$ term approximation to be
    \be 
    \label{ntermH}
    E(f,\Sigma_n)_H:=\inf_{S\in \Sigma_n}\|f-S\|_H.
    \ee 
    This  $n$-term approximation error from a dictionary is well studied.  A fundamental result for such $n$-term approximation is   the theorem of Maurey \cite{M} (see also \cite{B, Jones}).   Maurey's theorem says that for each $n\ge0$ and $f\in\cV(\cD_0)$ we have
    \be    
    \label{Mtheorem}
    \inf_{S_n\in \Sigma_n}\|f-S_n\|_H\le \|f\|_{\cV(\cD_0)}\delta n^{-1/2},\quad n\ge 1.
    \ee
    In fact, Maurey's theorem can be generalized beyond the setting of a Hilbert space to the class of type-$2$ Banach spaces (see \cite{SXSharp} for the application to non-linear dictionary approximation). This introduces an extra constant factor which depends upon the type-$2$ constant of the space.
We shall use this theorem going forward, but restrict ourselves to the Hilbert space setting.

  \section{Approximation in $\Omega=B^2$}
  \label{S:approxdictionary}

   In this section, we develop our  results  in the case   $\Omega=B^2$ where $B^2$ is the unit Euclidean ball in $\R^2$. Here, $\bar{Z}(\Omega) = S^1 \times [-1, 1]$. This will illustrate, in their simplest form, all of  the principles needed to treat   the more  general  case $\Omega=B^d$, $d\ge 2$.  The   treatment of $B^d$ is given
   in \S\ref{S:generald}  but with a significant increase in the level of technicality.

In this section, we let $\cD=\cD(\Omega)$ be the ReLU dictionary of atoms $\phi=\phi(\cdot;\xi,t)$,
$\xi\in S^1$ and $t\in [-1,1]$.
      Note that since $d=2$, the hyperplane $H_\phi$ associated to the atom $\phi$ is a line  and $L_\phi:=H_\phi\cap \Omega$ is a line segment whose length is $|L_\phi|=(1-t^2)^{1/2}$.
      We define the weight of this atom by
      \be 
      \label{aw2}
      w(\phi) = w(\phi(\cdot; \xi, t)) :=1-t,\quad t\in [-1, 1].
      \ee 
      It is easy to check that this weight is admissible since $\|\phi(\cdot; \xi,t)\|_{L_2(\Omega)}\approx (1-t)^{7/4}$ (see \eref{normphi}). We discuss where this weight comes from in \S\ref{S:generald} in the sequel.

    We first want to prove results on the linear approximation of the atoms $\phi$. Namely, for each $n=1,2,\dots$, we want to construct an $n$ dimensional linear  space $X_n$ which is good at approxating all of the atoms
    $\phi\in\cD(\Omega)$.  The linear space $X_n$ will be the span of $n$ well chosen atoms $\phi_j$, $j=1,\dots,n$, from $\cD(\Omega)$. The construction we give for $X_n$  is  a modification of ideas from \cite{SXSharp}. Our   analysis of the approximation error in approximating $\phi$ by the elements of $X_n$ is new in that it gives an improved error estimate when the support of $\phi$ is near the boundary of $\Omega$.

    To define the space $X_n:=\span\{\phi_1,\dots,\phi_n\}$, we want to choose the atoms $\phi_j$, $j=1,\dots,n$, to have as a special discrete distribution from $\cD$.
       In the case $d=2$,  these atoms are rather easy to describe geometrically as is given in the next paragraph.   When $d>2$, we will need more sophisticated arguments (see \S\ref{S:generald}).

 We fix $m\ge 4$ and let  $P=P_m$ be the set of points  
 \be  
 \mu_j=\mu_j(m):=(\cos \theta_{j}, \sin \theta_{j}), \quad  \theta_j=\theta_{j,m}:=\frac{2\pi j}m, \ j\in \Z.
 \ee  
 There are $m$ distinct points and $\mu_j=\mu_{j'}$ if $j$ and $j'$ are congruent modulo $m$, i.e., if  $j\equiv j'$.  These points are equally spaced on the circle. 
 
 Let $X_n$ be the linear space spanned by  the dictionary elements
 $\phi$ whose line segment $L_\phi$ has end points $\mu_i$ and $\mu_j$, $1\le i<j\le m$.  Notice that for each pair $i,j$ there are two such atoms.
Hence, the dimension of $X_n$ is $n:=m(m-1)$.  We also note that $X_n$ contains all linear functions on $\Omega$.
 
    Given $i,j\in\Z$, we define the distance between $i$ and $j$ by
   $$
    d(i,j) :=  \min\{|i'-j'|:\ i\equiv i',\ j\equiv j'\},
   $$
   i.e. to be the periodic distance between the indices $i$ and $j$.

   Let $\cL_{i,j}=\cL_{i,j}(m)$ be the set of all line segments $L$  whose end points $a,b$ are the points $(\cos \theta,\sin\theta)$ where $\theta\in[\theta_{i}, \theta_{i+1}]$ in the case of $a$ and $\theta\in[\theta_{j}, \theta_{j+1}]$ in the case of $b$.  We denote by $S_{i,j}=S_{i,j}(m)$ the union of all the line segments $L_\phi$  in $\cL_{i,j}$. %
   
    Note that the length $L_{i,j}$ and width $W_{i,j}$ of $S_{i,j}$ satisfy 
   \be
   \label{comparelength}
   |L_{ij}|\approx \frac{d(i,j)+1}{m},\quad |W_{ij}|\approx \frac{d(i,j)+1}{m^2},\quad 1\le i\le j\le m.
   \ee
   Here and later in this section, all constants of equivalence are absolute.
        It follows that the measure of $S_{i,j}$ satisfies
 \be
 \label{measS}
 |S_{i,j}|\lesssim    \frac{(d(i,j)+1)^2}{m^3},\quad 1\le i\le j\le m.
 \ee

  \begin{lemma}
  \label{L:AD}
  Suppose that $m\ge 4$ is an even integer, $n=m(m-1)$,  and   $\phi=\sigma(\cdot;\xi,t)$  is any dictionary element whose line segment $L_\phi$ is in $\cL_{i,j}=\cL_{i,j}(m)$ with $\mu_i\neq \mu_j$.  Then there is a function $g\in X_n$   such that
  
  \noindent
  {\rm (i)}  $\phi(x)=g(x),\ x\notin S_{i,j}$,
  
  \noindent
  {\rm (ii)} $ \|\phi-g\|_{L_\infty(\Omega)}\le C  \frac{d(i,j)}{m^2} $,
  with $C$ an absolute constant.
  
  \noindent
  {\rm (iii)}       
  $ \|\phi-g\|_{L_2(\Omega )}\le     Cw(\phi)n^{-3/4}$, 
    with $C$ an absolute constant.
\vskip .1in
\noindent
If $\phi\in\cL_{i,i}$ for some $i$, then there is a $g\in X_n$ such that statement {\rm (iii)} holds.

  \end{lemma}
    
  \noindent
  {\bf Proof:}   We first assume that  $0\le i<j\le m$.   Also, by reversing the roles of $i$ and $j$ if necessary, we can also assume that $j< m/2$.  Because of rotational symmetry we can assume that $i=0$, $i+1=1$, $0<j<m/2$. 
  Consider the linear function $\ell(x):= \xi \cdot x-t$.    Let the line segment $L_\phi=H_\phi \cap \Omega$ associated with $\phi$ be in
  $\cL_{i,j}$. Let $\mu_i=\mu_i(m)$, $i\in\Z$.  We use the following three functions $\phi_1,\phi_2,\phi_3$ in $X_n$ each of  whose line segments $L_{\phi_i}$ are contained in $\cL_{i,j}$.  Here, $L_{\phi_1}$ has endpoints $\mu_{i},\mu_{j+1}$, the second segment $L_{\phi_2}$ has end points $\mu_i,\mu_{j}$, and the third function $\phi_3$ has line segment $L_{\phi_3}$ with endpoints $\mu_{i+1},\mu_{j+1}$.  The orientation of these three atoms matches that of $\phi$.
  By this we mean that whenever $x\in\Omega$ is strictly outside $S_{i,j}$ and $\phi(x)>0$ then
  each of the functions $\phi_i$, $i=1,2,3$, will likewise be positive.  Similarly, if $x$ is strictly outside this strip and $\phi(x)=0$ the three functions $\phi_i$, $i=1,2,3$, will likewise vanish.
  
  Consider the three linear functions $\ell_j$, $j=1,2,3$, corresponding to these line segments.  That is, we have $\ell_i(x)=\xi_i'\cdot x-t_i'$  and $\phi_i(x)=\ell_i(x)_+$ with $\xi_i'\in S^1$ and $t_i'\in [-1,1]$. 
     Since these three linear functions are linearly independent, we can write
  \be
  \label{write1}
  \ell=c_1\ell_1+c_2\ell_2+c_3\ell_3.
  \ee
  Specifically, let $\zeta$ be the point where $\ell_2(\zeta)=\ell_3(\zeta) = 0$. Then,
    \be 
    \label{thecoeff}
    c_1= \frac{\ell(\zeta)}{\ell_1(\zeta)},\quad c_2= \frac{\ell(\mu_{j+1})}{\ell_2(\mu_{j+1})}, \quad c_3=\frac{\ell(\mu_{i})}{\ell_3(\mu_{i})}.
   \ee
   This follows by noting that with this choice \eqref{write1} holds at the affinely independent set of points $\zeta$, $\mu_{j+1}$ and $\mu_i$.
    
   We claim that
   \be
   \label{boundcoeff}
   |c_i| \le 1,\quad i=1,2,3.
   \ee
    Indeed, since the $\xi,\xi_i$ lie on the sphere it is clear that $|\ell_i(x)| = d(x,L_{\phi_i})$ and $|\ell(x)| = d(x,L_\phi)$ for any $x\in \mathbb{R}^2$ (here $d(x,L)$ denotes the distance from the point $x$ to the line $L$). We will show that 
   \be
   \label{toshow1}
   d(\mu_i,L_\phi) \leq d(\mu_i,L_{\phi_3}),
   \ee
   which implies $|c_3| \leq 1$. 
   A completely analogous argument shows that $|c_2| \leq 1$.
   
    For the proof of \eref{toshow1}, we assume that $j>i+1$.  If $j=i+1$, a similar argument applies (which we leave to the reader). Consider the trapezoid whose vertices are $\mu_i, \mu_{i+1}, \mu_j, \mu_{j+1}$ and let $T$ denote its interior.  
   Let $\bar{\mu}_i$ denote the orthogonal projection of $\mu_i$ onto the line $L_{\phi_3}$.  The angle formed by the vertices $\mu_{j+1}, \mu_{i+1},\mu_i$ is larger than or equal to $\pi/2$.  This means that  $\bar{\mu}_i$ lies either on or outside of the circle.  By the defining property of $L_\phi$ this line must intersect the segment   $[\mu_i,\bar{\mu}_i]$.  Therefore, $d(\mu_i,L_\phi) \leq d(\mu_i,L_{\phi_3})$ which proves \eref{toshow1} as desired.

    Next, we consider bounding $|c_1|$. The line segments $[\mu_i,\mu_{j+1}]$ and $[\mu_{i+1},\mu_j]$ are parallel, and the intersection point $\zeta$ lies on the perpendicular line $L_p$ connecting the midpoints of these two line segments.  Moreover,     the lengths of these segments satisfy $l([\mu_i,\mu_{j+1}]) > l([\mu_{i+1},\mu_j])$. This means that the distance from $\zeta$ to $L_{\phi_1}$ is greater than the distance to the parallel line segment $[\mu_{i+1},\mu_j]$. Finally, since $L_\phi\in \cL_{i,j}$, $L_\phi$ must intersect $L_p$, which implies that $d(\zeta,L_\phi) \leq d(\zeta,L_{\phi_1})$. This means that $|c_1| \leq 1$ as claimed.

    Now, consider the function 
   \be 
   \label{defg}
   g:=c_1\phi_1+c_2\phi_2+c_3\phi_3,
   \ee 
   which is in the linear space $X_n$.  This function agrees with $\phi$ outside $S_{i,j}$ so that
   (i) is satisfied. Each of the functions $\phi$ and $g$ have $L_\infty(S_{i,j})$ norm not exceeding the width $W_{ij} \leq C\frac{d(i,j)}{m^2}$ (they are $1$-Lipschitz and vanish on one edge) and so the  upper bound in (ii) follows.
   The function  $\phi-g$ is supported on $S_{i,j}$ and we have 
   \be
  \|\phi-g\|_{L_2(\Omega )}\le  \|\phi - g\|_{L_\infty(\Omega)} |S_{i,j}|^{1/2}\le C d(i,j)^{2} m^{-2-3/2} \le C|L_{ij}|^{2}m^{-1-1/2}.
  \label{phipnorm}
\ee
      Note that in this calculation we have use that $d(i,j) \approx (d(i,j) + 1)$. Since $d(i,j) > 1$, we easily see that $|L_{ij}| \approx |L_\phi| = w(\phi)$, which verifies (iii).
      
      Finally, if $L_\phi $ is in $\cL_{i,j}$ with $d(i,j) \leq 1$ then the conclusion follows   in the same way we proved \eref{phipnorm} by taking either $v=0$ or $v = w\cdot x + b$ to be linear function which matches the  linear part of $\phi$. 
          \hfill $\Box$

    \subsection{The approximation theorem}
    \label{SS:d2}
    
    Throughout this section $E_n(f):= E_n(f)_{L_2(\Omega)}$, $n\ge 1$ for any $f\in L_2(\Omega)$. 
    We can now state  the main theorem  to be proved in  this section.
        \begin{theorem}
   \label{T:d2}
   Let    $\Omega=B^2$ and $w(\phi)$, $\phi\in\cD(\Omega)$, be defined by \eref{aw2}.  Then for any $f\in\cV_w$,  we have
   \be
   \label{Ta1}
   E_{n}(f) \le C \|f\|_{\cV_w(\Omega)} n^{- \frac {5}{4}}, \quad n\ge 1,
   \ee
   where $C$ is an absolute constant.     \end{theorem}
   \vskip .1in
   \noindent
   {\bf Proof:}    Since $\Sigma_n\subset \Sigma_{n+1}$, $n\ge 0$,  it is enough to prove the theorem for any $n=m(m-1)$ with $m\ge 4$ an even integer.  This means that we can apply Lemma \ref{L:AD}.    It is enough to prove the theorem for any function  $f$ from $U(\cV_w(\Omega))$. According to the definition of $\cV_w(\Omega)$, for $N$ sufficiently large, there is an $S\in\Sigma_N$ with $S=\sum_{j=1}^Na_j\phi_j$  such that
   \be
   \label{T11}
   \|f-S\|_{L_2(\Omega)}\le n^{-5/4}\quad {\rm and} \quad \sum_{j=1}^N w(\phi_j) |a_j|\le 1.
   \ee
   For each $j$, let 
    $g_j\in X_n$ approximate the function $\phi_j$ appearing in the representation of $S$ according to (iii) of Lemma \ref{L:AD}.
    That is, we have
    \be
    \label{wehave}
    \|\phi_j-g_j\|_{L_2(\Omega)}\le C_0 w(\phi_j)  n^{-3/4},
    \ee
    with $C_0$ an absolute constant.  The function $g:=\sum_{j=1}^N a_jg_j$ is in $X_n$ and hence in $\Sigma_n$.  We write
    \be
    \label{writedecomp}
    f= f-S+h+g,\quad h:=S-g.
    \ee
      Therefore,
      \be
      \label{T1}
      E_{3n}(f)\le  n^{-5/4} +E_{2n}(h).
      \ee
      We want to bound $E_{2n}(h)$.   We have $h=\sum_{j=1}^N a_j[\phi_j-v_j]$.  
      We consider the dictionary $\cD'=\{\psi_j\}_{j=1}^N$ with $\psi_j:= w(\phi_j)^{-1}(\phi_j-g_j)$.    According to \eref{T11} and \eref{wehave},  each $\psi_j$ has $L_2(\Omega)$ norm at most $C_0n^{-3/4}$ and $h=\sum_{j=1}^Nc_j'\psi_j$ with $\sum_{j=1}^N|c_j'|\le 1$.  It follows from  Maurey's theorem (see \eref{Mtheorem}) that $h$ can be approximated by a sum $T$ of $n$ terms from the dictionary $\cD'$ with error
      \be
      \label{T12}
      \|h-T\|_{L_2(\Omega)} \le C n^{-3/4}n^{-1/2}=Cn^{-5/4},
      \ee
      with $C$ an absolute constant.
         The function $T $ is a sum of at most $2n$ terms from the original dictionary $\cD$.  Hence,
      \be
      \label{T13}
      E_{2n}(h) \le C n^{-5/4}.
      \ee
    If we place this inequality back into \eref{T1}, we obtain 
   \be
      \label{T14}
      E_{3n}(f)\le  [1+C]n^{-5/4}
      \ee
  and the theorem follows.\hfill $\Box$

     We close this section with two remarks that clarify Theorem \ref{T:d2}.
     
     \begin{remark}
         \label{R:emphasize}
We emphasize that Theorem \ref{T:d2} is an improvement on the
      known theorem that any $f\in\cV(\cD)$ satisfies $E_n(f)\le Cn^{-5/4}$ because the weighted
      variation space $\cV_w$ is strictly larger than the standard variation space $\cV$.
         
     \end{remark}

      \begin{remark}
         \label{R:generalize}
While Theorem \ref{T:d2} only applies to the approximation rate $O(n^{-\alpha})$, when $\alpha=5/4$, there is a standard technique to obtain results for more general rates $O(n^{-\beta})$, for any $\beta\le 5/4$,  by considering the interpolation spaces between $L_2(\Omega)$ and $\cV_w(\Omega)$ as is explained in
\cite{DP} and \cite{BCDD}.  This approach gives new sufficient conditions for membership in $\cA^\beta$.  This remark also applies to later results in this paper. We do not elaborate further on this point.
         
     \end{remark}

   \subsection{Weighted variation spaces for $\Omega=Q^2$}
   \label{SS:Q2}
    Although we do not formulate a general result, it will be clear that the techniques of this paper can be generalized to any convex domain $\Omega$.  In this section, we want to point out what such a result is for $Q^2:=[-1,1]^2$  since this will allow us to see the effect of the geometry of $\Omega$.    So, in going further in this section, we
    take $\Omega=Q^2$.  
    
    If $\phi$ is a ReLU atom, then the line segment $L_\phi$ relative to $\Omega$ is   $H_\phi\cap\Omega$.  The length
   $|L_\phi|$ can now be large even if $L_\phi$ is close to the boundary of $\Omega$,  for example when $L_\phi$ is parallel to 
   one of the sides of $\Omega$.  In other words, many fewer atoms $\phi$ will have small $|L_\phi|$.
   
   Let us sketch how the results and analysis for approximating general atoms $\phi$ given in \S \ref{SS:d2} for $B^2$,  changes in this case.   We now take a set of $m\sim \sqrt{n}$ equally spaced points on the boundary of $Q^2$.  We can associate
   each $\phi$ to a $\cL_{i,j}$ similar to the case of $B^2$ and create a linear space $X_n$ of dimension $m^2\sim n$ as before.  Now the analogue of Lemma \ref{L:AD}  says that any dictionary element $\phi$ can be approximated by an element of $g\in X_n$ to an accuracy  (corresponding to (iii) in that lemma)
   \be
   \label{a2}
   \|\phi-g\|_{L_2(\Omega)} \le C_0 m^{-1} [|L_\phi| m^{-1}]^{1/2} =C_0m^{-3/2} |L_\phi|^{1/2}.
   \ee
   Here the factor $m^{-1}$ reflects the $L_\infty$ error and the bracketed factor is the measure of the support where $\phi$ and $g$
   differ.    
   
   Given the above calculations, we define $w(\phi):=|L_\phi|^{1/2}$  as the weight of the atom $\phi$ and use this weight to define
   $\cV_w(Q^2)$. The proof of Theorem \ref{T:d2}   now gives
     \begin{theorem}
   \label{T:Q2}
   Let  $d=2$ and  $\Omega=Q^2$ and define $w(\phi):=|L_\phi|^{1/2}$.  Then for any $f\in \cV_w$,  we have
   \be
   E_{n}(f) \le C \|f\|_{\cV_w(Q^2)}n^{- \frac {5}{4}}, \quad n\ge 1,
   \ee
   where $C$ is an absolute constant.     \end{theorem} 
  
    \section{Approximation in $L_2(B^d)$}
   \label{S:generald}

   We turn now to the case of approximation on the domain $\Omega=B^d$, $d>2$,  i.e.,  $\Omega$ is the unit  Euclidean ball
   of $\R^d$.   
       Recall that each atom $\phi=\phi(\cdot;\xi,t)$, satisfies $(\xi, t) \in \bar{Z}(\Omega) = S^{d-1} \times [-1,1]$.  To each atom, we assign the special weight
   \be  
   \label{defweightd}
   w(\xi,t):= w^*(\xi,t) := (1-t)^{\frac{1}{2}+\frac{d}{4} }.
   \ee
    From \eref{normphi}, we see that this weight is admissible for $\Omega$. Thus, $w$ is taken to be given by \eref{defweightd}
    throughout this section.
        One can ask where the particular form of the weight comes from. It arises due to the norm of the atoms in $L_2$ and the smoothness of the parameterization of the atoms. The effects of these two ingredients will become clear in the details of the proof.

    We recall the variation space $\cV_w$ introduced and studied in \S \ref{S:variation}. 
The main result of this paper is the following theorem
\begin{theorem}
    \label{T:B^d}
    If $f\in \cV_w=\cV_w(\Omega)$ then
    \be  
    \label{T21}
    E_n(f):=E_n(f)_{L_2(\Omega)}\le C\|f\|_{\cV_w}  n^{-\frac{1}{2}-\frac{3}{2d}}, \quad n\ge 1,
    \ee  
    where $C$ depends only on $d$.
\end{theorem}
\noindent
Notice that this theorem gives a stronger result than the previously known results on approximation by shallow neural networks with ReLU activation.  Indeed, although the approximation rate 
$O( n^{-\frac{1}{2}-\frac{3}{2d}})$ is the same as known whenever $f\in\cV$, the assumption of   membership in $\cV_w$
is a strictly weaker assumption than the membership in the traditional  variation space $\cV$.
       
   The remainder of this section is devoted to proving Theorem \ref{T:B^d}.  The proof is similar, in spirit, to the case $d=2$ which was given in Theorem \ref{T:d2}, but it is quite a bit more technical.
   Our first goal is to construct
    certain  linear  spaces $X_n$ of dimension at most $n$, which can be used to effectively approximate general ReLU atoms.
    The space $X_n$ will be the span of at most $n$  well chosen atoms from $\cD(\Omega)$.  The choice of the atoms used to define $X_n$ would intuitively be gotten by    discretizing the unit Euclidean sphere $S^{d-1}$ with $m^{d-1}$ uniformly spaced vectors and then  
    and to discretize the offsets in $T=[-1,1]$ with $m$ points.  Here $m$ is chosen so that $n\approx m^d$.   The discretization of $T$ will not be uniform but instead  will be be done in such a way that atoms whose support is small, i.e., atoms whose associated hyperplane lies near the boundary $S^{d-1}$ of $B^d$ will be very well approximated.
    
    Since there is no natural discretization of $S^{d-1}$, when $d>2$,  we proceed as follows. 
    Let $Q:=Q^d:=[-1,1]^d$ and $F$ be a face of $Q$. Each face $F$ is gotten by setting one of the coordinates, say, coordinate $i$,  equal to either $+1$ or $-1$.  Given one of these faces $F$, 
    we shall use dyadic partitions of $F$ into $d-1$ dimensional cubes of side length $2^{-k}$.
    We let $V_k(F)$ be the set of all vertices of this partition.  
     Thus, the cardinality of $V_k(F)$ is $(2^k+1)^{d-1}$.  We define $V_k$ to be the union    of all of the sets $V_k(F)$ as $F$ runs through the $2d$ faces of $Q$.  This gives a discrete set of points on the boundary of $Q$ with $\ell_\infty$ spacing $2^{-k}$.   To obtain our discrete set of points on $S^{d-1}$, we simply renormalize.  Namely,
    \be  
    \label{Wm}
    W_k:=\left\{\xi=\frac{\bar \xi}{\|\bar \xi\|}:\ \bar \xi\in V_k\right\}
    \ee
    gives a set of points on the boundary of $B^d$ that are quasi-uniformly spaced
     in the sense that
   \be
   \label{qu}
    c_0 2^{-k}\le \dist (\xi_i, W_k\setminus \{\xi_i\}) \le C_02^{-k},
   \ee
   where the constants\footnote{In this paper, all constants depend only on $d$ and may change from line to line.  We use $c$ for small constants and $C$ for large constants, sometimes with subscripts.} depend only on $d$.  After adjusting for redundancy, we see that the cardinality of $W_k$ is   $2d(2^k)^{d-1}$.  It is important to note that
   \be  
   \label{nesting}
   V_k\subset V_{k+1} \quad {\rm and} \quad W_k\subset W_{k+1},\quad k\ge 1.
   \ee 
   
   We also want to discretize the offsets $t$.  For this, we take  
   \be  
   \label{Tm}
   T_m:=\{-1<t_1<\dots   < t_{2m}=1\}.
   \ee
   We take the first $m$ of these to be equally spaced in $[-1,0]$, i.e., $t_j:=-1+j/m$, $j=1,\dots,m$.  For the remainder of these points, we take

\be  
\label{discretet}
     t_{j+m}:= \cos \frac{\pi (m-j)}{2m}=:\cos \theta_{j,m},\quad j=1,\dots, m.
     \ee
   Notice that the points in $T_m$ have a finer spacing near one.  Concerning this spacing, in going further we will use   the fact that for each  $m<j<2m-1$ and $t\in [t_j,t_{j+1}]$ we have
   \be  
   \label{factspacing}
 \frac{ \pi\sqrt{1-t_{j+1}^2}}{2m}\le  |t_{j+1}-t_{j}|\le \frac{\pi \sqrt{1-t_j^2}}{2m}  \quad  {\rm and} \quad \sqrt{1-t_{j}^2}\le   2 \sqrt{1-t_{j+1}^2}\le 2 \sqrt{1-t^2}.
   \ee 
   To prove this, we note that for fixed $j=i+m$, $1\le i<m$, we have
   $$|t_{j+1}-t_j|= \frac{\pi}{2m} \sin\zeta =\frac{\pi}{2m}\sqrt{1-\cos^2 \zeta} $$ 
   where $\zeta\in [\theta_{i+1,m},\theta_{i,m}]$.  This gives the first inequalities in \eref{factspacing}.
   The second inequalities are proved similarly.
   The inequalities in \eref{factspacing} show that  for any given $t\in [t_j,t_{j+1}]$, $j\le 2m-2$, we have $|t_{j+1}-t_j|\approx  \sqrt{1-t^2}/m$ 
    with absolute constants in this comparison.  We shall use this fact repeatedly.
    
    We now want to define the linear space $X_n$.  Consider the set of atoms given by  
   \be 
   \label{phim}
   \Phi_{k,m}:=\{\phi(\cdot;\xi,t),\ \xi \in W_k,\ t\in T_m\}.
   \ee 
   This is a set of at most $ 4d m(2^k)^{d-1}$ ReLU atoms.  We  choose $k$ as the largest integer   such that $ 4d 2^k (2^k)^{d-1}\le n$ and then take $m=2^k$.
   Then, $\Phi_n:= \Phi_{k,m}$ is a set of at most $n$ atoms.
   We define 
   $X_n$ 
 as the linear space  
 \be  
 \label{Vatoms}
 X_n:=\span (\Phi_n),
 \ee 
 Then, $X_n$ is a linear  space of  dimension  at most  $n$.

 We caution the reader that for the remainder of this paper, the integer $n$ is always taken of the form $n=4dm^d$, where $m=2^k$.
 It is enough to prove our approximation results for these $n$.

We now  proceed to show that any atom $\phi:=\phi(\cdot;\xi,t)$ from $\cD(\Omega)$  can be well approximated by
an element of the linear space $X_n$. We fix $\xi, t$ and $n$.  The approximation result we prove
is given in the following theorem.

\begin{theorem}
    \label{T:approxphi}
    For any $(\xi,t)\in\bar Z(\Omega)=S^{d-1}\times [-1,1]$, there is an element $g=g_\phi\in X_n$ such that
    \be 
    \label{Taphi}
    \|\phi-g\|_{L_2(\Omega)}\le C w(\phi)n^{-\frac{3}{2d}},
    \ee 
    with the constant $C$ depending only on 
    $d$.
\end{theorem}
The proof of this theorem is a bit technical and given in the next subsection.  After proving this 
theorem we prove Theorem \ref{T:B^d}. In the constructions given below there are two important constants $A$ and $L$ which depend only on $d$.  It will be useful to the reader if we explain their role and their definition. To prove Theorem \ref{T:approxphi}, we are presented with an atom
$\phi(x) =(\xi\cdot x-t)_+$ and need to construct an element $g\in X_n$ that approximates $\phi$ to the given accuracy.  From the definition of $X_n$, the function $g$ will take the form
\be 
\label{gform}
g=\sum_{j=1}^na_j\phi_j,
\ee 
where the $\phi_j$ are the atoms $\phi_j(x)=(\xi_j\cdot x-\tau_j)_+$ used to define $X_n$, where $\tau_j = t_{i(j)}$. The function $g$ that we construct to provide the approximation will agree with $\phi$
except on a certain set of small measure.  The only atoms active
in the definition of $g$, i.e., which have nonzero coefficients, will satisfy $\|\xi-\xi_j\|\le \frac{A}{m}$,
with $A\ge 2$ a fixed  integer constant depending only on $d$.  The size of $A$ is determined by the proof of Lemma \ref{L:decomega}
which is formulated later in this section and then proved  in the Appendix. Hence, in going forward, we can consider   $d$
arbitrary but fixed and $A$ depending only on $d$ to be fixed as well.

The constant $L$ is an integer which  is chosen in the proof of Lemma \ref{L:ti+}.  We will only have to consider values of $t$ such that $t\le t_{2m-L}$.  This restriction can be applied on $t$ because of the following remark.   

\begin{remark}
    \label{R:proof}
     Let us note and record the following: 
    
    \noindent {\rm (i)} If $t\ge t_{2m}$ or even $t\ge t_{m'}$ with $m'=2m- L$ with $L$ a fixed integer, then
    for any atom $\phi(\cdot;\xi,t)$ the statement \eref{Taphi} holds by simply taking $g=0$ and using the
    estimate \eref{normphi} for the norm of the atom.

\noindent {\rm (ii)} If $t\le C<1$ with $C$ fixed then the weight $w(\xi,t)\ge c$.  In this case, the existence of a space
$X_n$ spanned by $n$ atoms that provides the estimate \eref{Taphi} was given in \cite{SXSharp}.  While our space $X_n$ is defined differently  (we use a different discretization of the offsets $t$), the proof in this case is simpler and we exclude this case going forward.

\noindent {\rm (iii)}  If $\xi$ is one of the discrete vectors from $W_k$, then the proof of the existence of a $g$ for which \eref{Taphi} is quite simple. Indeed, one can take $g= a\phi(\cdot;\xi,t_i) +(1-a)\phi(\cdot;\xi, t_{i+1})$ where $t_i$ is the closest discrete offset to $t$ and $a$ is chosen so that $at_i+(1-a)t_{i+1}=t$.

In the proof of \eref{Taphi} we only need to provide a proof in the case that none of the special cases {\rm (i-iii)} holds.

\end{remark}

\subsection{The proof of Theorem \ref{T:approxphi}}
\label{SS:Proofapproxphi}
Obviously, we only need to prove the theorem for $m$ sufficiently large, say $m\ge m^*$ where $m^*$ depends on $d$.  The integer $m^*$ will be specified as we go along. Because of Remark \ref{R:proof} we only need to prove the theorem in the case $1/2<t\le t_{m-L}$, where $L$ is a fixed integer depending only on $d$. Again, we shall specify $L$ as we proceed in the proof.  Similarly, we can assume $\xi$ is not in $W_k$. 
We fix such an $\xi\in S^{d-1}$ and such a $t$ throughout this subsection.

            We define
\be  
\label{defH}
H^+:=\{x:\ \xi\cdot x\ge 
t\} \quad  {\rm and} \quad H^-:=\{x:\ \xi\cdot x < t\}.
\ee
So $\phi$ is identically zero on $H^-$ and the linear function $\xi\cdot x -t$ on $H^+$.
For  any  one of the vectors $\xi_i$ appearing in the set $W_k$   and  any given a $t_j\in T_m$, we similarly define    
\be  
\label{hyper}
H_j^+:=H_j(\xi_i):=\{x\in\Omega: \xi_i\cdot x \ge t_j\},\quad H_j^-:=H_j^-(\xi_i):=\{x\in\Omega: \xi_i\cdot x<  t_j\}.
\ee

Given $\xi_i$, we want to choose a $t_j$ with $j\le 2m-1$ (depending on $i$) that is close to $t$ and so that $H_j^+$ is a subset of $H^+$.
This is always possible whenever $\|\xi-\xi_i\|\le A/m$ and  $t\le t_{m-L}$ and $L$ is sufficiently large (depending only on $A$).
One such choice for $t_j$ is to take
\be  
\label{appt1}
 t_i^+:=t^+(\xi_i):=\min \{t_j\in T_m:  H_j^+\subset H^+\}.
\ee

 If $t_i^+=t_j$,  we let 
 \be
 \label{deftildeT}
 \tilde t_i:= t_{j+1}.
 \ee
 Then, we will also have $H^+_{j+1}\subset H^+$.

 We now proceed to proving Theorem \ref{T:approxphi}.  We begin by  recalling  the following fact.
\begin{lemma}
    \label{L:ip}
    If $\xi,\xi'\in S^{d-1}$ with $\|\xi-\xi'\|=\delta$, then we have
    \be 
    \label{Lip}
    \xi\cdot\xi'=1-\delta^2/2.
    \ee
\end{lemma}
\vskip .1in
\noindent{\bf Proof:} By rotation, we can assume $\xi=e_1=(1,0,\dots,0)$ and $\xi'=\alpha e_1+\eta$ where $\eta$ is orthogonal to $e_1$ and $\|\eta\|^2=1-\alpha^2$.  Therefore,
$$
\delta^2= (1-\alpha)^2+\|\eta\|^2 = 2-2\alpha =2-2\xi\cdot\xi' $$
 and so \eref{Lip} follows. \hfill $\Box$

The last lemma allows us to compare $t_i^+$ with $t$.
\begin{lemma}
     \label{L:ti+}
    Given the integer $A$, we define 
    \be 
    \label{defL}
    L:=(A+1)^2=A^2+2A+1.
    \ee 
    If $m^*$ is sufficiently large, depending only on $d$ and $m\ge m^*$, then  whenever  $t\in [1/2,t_{2m-L}]$  and $ \|\xi-\xi_i\|\le \frac{A}{m}$,  then $t_i^+$ and $\tilde t_i$ are well defined, and  we have
     \be 
     \label{Ltin}
     t\le t_i^+ \le \tilde t_i\le  t+\frac{C_1}{m}\sqrt{1-{t}^2}, 
     \ee 
     where $C_1 $ depends only on $d$.
\end{lemma}
\vskip .1in
\noindent 
{\bf Proof:}  
     Consider first the existence of $t_i^+, \tilde t_i$.   It is enough to show that if $t\le t_{2m-L}$, and $\xi_i$ satisfies $\|\xi-\xi_i\|\le \frac{A}{m}$, then there is a $j\le 2 m-2$ such that $H_j^+\subset H^+$. Suppose that $j$ is such that $t_{j}\ge t$ but $H_{j}^+$ is not contained in $H^+$.   Then,    there is an $x=t_{j}\xi_i+\eta$
     with $\eta$ orthogonal to $\xi_i$ and $\|\eta\|\le \sqrt{1-t_{j}^2}$ and
     $$
     x\cdot\xi = t_{j}\xi\cdot\xi_i + \eta\cdot(\xi-\xi_i)< t .
     $$
      From Lemma \ref{L:ip}, we know that $\xi\cdot \xi_j\ge 1-\frac{A^2}{m^2}$ and so we must have
      
    \be 
    \label{Ltin1}
       \left(1-\frac{A^2}{m^2}\right) t_{j} \le t+ \|\xi-\xi_i\|\sqrt{1-t_{j}^2}\le  t+ \frac{A}{m}\sqrt{1-t_j^2}.
     \ee  
      That is, we must have
      \be 
    \label{Ltin11}
     t_{j}\le  t+ \frac{A}{m}\sqrt{1-t_j^2}+\frac{A^2}{m^2}\le  t_{2m-L}+ \frac{A}{m}\sqrt{1-t_j^2}+\frac{A^2}{m^2}.
     \ee  
If we write $j=2m-k$, and use the definition of the $t_j$ (see \eref{discretet}) we can rewrite \eref{Ltin11}
as
\be 
\label{Ltin12}
     \cos \frac{\pi k}{2m} - \cos \frac{\pi L}{2m}  \le     \frac{A}{m}\sin \frac{\pi k}{2m}+\frac{A^2}{m^2}
     \le \frac{\pi A k}{2m^2} +\frac{A^2}{m^2}\le \frac{A^2+2Ak}{m^2}.
     \ee 
The left side of \eref{Ltin12} is larger than $\frac{k(L-k)}{m^2}$ and so we see with the above definition of $L$,
\eref{Ltin12} is violated when $k=2$.   This proves that $t_i^+$ and $\tilde t_i$ are well defined.

We turn now to proving \eref{Ltin}.
      First note that if there is  $j\in\{m+1,\dots,2m\}$   such that $H_j^+\subset H^+$,
 then  we must have
     $t_j\xi_i\cdot \xi\ge t$ which gives the left inequality in
     \eref{Ltin}.   To prove the right  inequality in \eref{Ltin}, we let $t_i^+=t_j$, $\tilde t_i=t_{j+1}$.  It follows
     from the minimality in the definition of $t_i^+$ that we must have $H_{j-1}^+$ is not contained in $H^+$.  Thus, the inequality \eref{Ltin1} holds with $j$ replaced by $j-1$. This gives
      \be 
    \label{Ltin2}
      \left(1-\frac{A^2}{m^2}\right)  t_{j-1}  \le t+ \|\xi-\xi_i\|\sqrt{1-t_{j-1}^2}\le  t+  \frac{2A}{m}\sqrt{1-{t_i^+}^2},
     \ee  
   where the last inequality uses \eref{factspacing}. 
      From \eref{factspacing}, we also have
      $$ 
       t_j\le t_{j-1}+ (t_j-t_{j-1}) \le  t_{j-1}+ \frac{\pi}{m}\sqrt{1-t_j^2}.
      $$
        If we multiply both sides of this last inequality by $(1-\frac{A^2}{m^2}) $ and use \eref{Ltin2} we obtain 
$$   \left(1-\frac{A^2}{m^2}\right)  t_i^+ \le  t+\frac{2(A+2)}{m}\sqrt{1-t^2},$$
      where we used that $t_i^+\ge t$ .    We also have $\tilde t_i\le t_i^++ \frac{\pi}{2m}\sqrt{1-t^2}$ because of \eref{factspacing}.  When these facts are used in the last inequality we   obtain the right inequality       in \eref{Ltin}.\hfill $\Box$
    
  Now consider any $\xi\in S^{d-1}$ and define
  \be 
  \label{defWomega}
  W_k(\xi):=\left\{\xi_i\in W_k: \ \|\xi-\xi_i\|\le \frac{A}{m}\right\}\quad{\rm and} \quad t^+:=t^+(\xi):=\max_{\xi_i\in W_k(\xi)}t_i^+.
  \ee 
  We can write $t_i^+=t_j$ for some $j\le 2m-1$ and define $\tilde t:=\tilde t(\xi):=t_{j+1}$. From the previous lemma, we know that
   
  \be 
     \label{Ltin3}
     t\le t^+ \le \tilde t\le  t+\frac{C_1}{m}\sqrt{1-{t}^2}, 
     \ee 
     where $C_1 $ depends only on $d$.
  
For the construction of $g$, we use the following  lemma.  

\begin{lemma}
\label{L:decomega}
There is a constant $m^*$ depending only on $d$   such that the following holds.
Given any $m\ge m^*$ and any $\xi\in S^{d-1}$ and any $ 1/2\le t\le t_{2m-L}$, there exists  $(a_i^+) $, $(\tilde a_i) $ such that
\vskip .1in
\noindent 
{\rm(i)}\quad 
       $ \xi = \sum_{\xi_i\in W_k(\xi)} a^+_i\xi_i + \sum_{\xi_i\in W_k(\xi)} \tilde a_i\xi_i $,
  \vskip .1in

\noindent 
{\rm(ii)}\quad     
     $t^+\sum_{\xi_i\in W_k(\xi)} a^+_i + \tilde t \sum_{\xi_i\in W_k(\xi)} \tilde a_i  = t$,
     \vskip .1in
     \noindent 
{\rm(iii)}\quad     
      $\sum_{\xi_i\in W_k(\xi)} |a^+_i| + \sum_{\xi_i\in W_k(\xi)} |\tilde a_i| \leq C_1$, 
  where $C_1$ depends only on $d$.   

\end{lemma}
\noindent
The proof of this lemma is technical and so we place it in the Appendix so as not to interrupt the flow of the proof of  Theorem \ref{T:approxphi}.

We define the following function $g$ which will be used to approximate $\phi=\phi(\cdot;\xi,t)$ in the case
$1/2\le t\le t_{2m-L}$
\be\label{defv}
g(x):=\sum_{\xi_i\in W_k(\xi)}[ a^+_i (\xi_i\cdot x -t^+)_+ +   \tilde a_i(\xi_i\cdot x-\tilde t)_+]
\ee
where the coefficients   come from Lemma \ref{L:decomega}.
The functions appearing in the representation of $g$ are all in $X_n$ and therefore $g$ is also in $X_n$.
 From Lemma \ref{L:decomega},  we obtain
\be\label{yes}
\phi(x)-g(x)=\left[ \sum_{\xi_i\in W_k(\xi)} a^+_i (\xi_i\cdot x -t^+) +  \tilde a_i(\xi_i\cdot x-\tilde t)\right]_+  -\sum_{\xi_i\in W_k(\xi)} [a^+_i (\xi_i\cdot x -t^+)_+ +  \tilde a_i(\xi_i\cdot x-\tilde t)_+]
\ee

Before bounding the error in approximating $\phi$ by $g$, let us make some remarks to motivate the idea
of how to estimate this error. Notice that if $x\in \Omega$ is such that $\xi_i\cdot x\ge \tilde t$ for all $\xi_i\in W_k(\xi)$,  then $x$ is also in $H^+$ and so
  $g(x)=\phi(x)$.  Similarly, if 
$x\in H^-$ then   $\phi(x)=g(x)=0$.  This means that the only points $x\in \Omega$ where
$\phi(x)\neq g(x)$ must be in one of the sets
\be  
\label{must}
\tilde \Omega_i:=\{x\in\Omega: \xi\cdot x> t,\ \xi_i\cdot x\le   \tilde t\}.
\ee  
We will now proceed to bound the measure of each of these sets and also bound the error $|\phi(x)-g(x)|$ on each of these sets.

\begin{lemma}
 \label{L:error}
There are constants $C$ and $m^*$ depending only on $d$ such that the following holds for $m\ge m^*$ and any $1/2\le t\le t_{2m-L}$:

\noindent
{\rm (i)} If $x\in \tilde \Omega:= \bigcup_{\xi_i\in W_k(\xi)} \tilde \Omega_i$, then 
 \be  
 \label{Le1}
 |\phi(x)-g(x)|\le C\sqrt{1-t^2}/m.
 \ee 
 \noindent{\rm (ii)} The measure of $\tilde \Omega$ is bounded by
 \be 
 \label{Le2}
  |\tilde \Omega|_d\le C (1-t^2)^{d/2}/m.
 \ee

   \end{lemma}
 \vskip .1in
\noindent
{\bf Proof:}  
From Lemma \ref{L:ti+} we have that

\be  
\label{boundti2}
t\le  t^+\le \tilde t \le  t+C_1\frac{\sqrt{1-t^2}}{m},\quad i=1,\dots,M
\ee 
where $C_1$ depends only on $d$.
Now, for any fixed $i$ consider any $x\in \tilde \Omega_i$.  Our   goal is to estimate the distance from $x$ to the hyperplane $H:=\{z:\ z\cdot \xi=t\}$.  From the definition of $W_k(\xi)$, we have
$\|\xi-\xi_i\|\le  A/m$.  Since $\alpha:=x\cdot\xi>t$, we can write  
   \be  
   \label{repx} x= \alpha   \xi+  \eta, 
   \ee  
   where   $ \eta$ is orthogonal to $  \xi$ and  
   \be  
   \label{eta}
   \|  \eta\|\le \sqrt{1-\alpha^2}\le \sqrt{1-t^2} .
   \ee
   We want to show that $\alpha$ cannot be too large.  Since $x\in \tilde \Omega_i$, we know that 
   \be  
   \label{proj}
   (x\cdot\xi_i)=\alpha\xi\cdot \xi_i + \eta\cdot\xi_i \le \tilde t ,
   \ee  
   and
   \be  
   \label{note}
   |  \eta\cdot\xi_i|=| \eta\cdot (\xi_i- \xi)|\le A\frac{ \sqrt{1-  t ^2}}{m}.
   \ee  
Using   the inequality $\xi\cdot\xi_i\ge 1-A^2/m^2$ (see Lemma \ref{L:ip}) and \eref{note} back in \eref{proj}, we obtain
\be  
\label{gives1}
  (1-A^2m^{-2})\alpha \le \tilde t + A\frac{ \sqrt{1-  t ^2}}{m}\le  t+C_1\frac{\sqrt{1-t^2}}{m} +A\frac{ \sqrt{1-  t ^2}}{m} , 
 \ee 
 where the last inequality used \eref{boundti2}.  Going further, we note that since by assumption
 $t\le t_{2m-L}$, we have
 $$(1-A^2m^{-2})^{-1}\le 1+2Am^{-2}\le 1+2A\frac{\sqrt{1-t^2}}{m}.$$
 Using this estimate back in \eref{gives1}, we arrive at
 \be  
 \label{arriveat}
 \alpha \le t+C_2\frac{\sqrt{1-t^2}}{m},
 \ee
 with $C_2$ depending only on $d$.
 It is important to note that this bound is independent of $i$.  Therefore any point $x$ in $\tilde \Omega$
 can be written as $x=\alpha\xi+\eta$ where $\alpha $ satisfies \eref{arriveat} and $\|\eta\|\le \sqrt{1-t^2}$.  The measure of the set of such $\eta$ is $\le C_3[\sqrt{1-t^2}]^{d-1}$. Hence, we have proven (ii).  
 
  The inequality \eref{arriveat} also shows that  
 \be 
 \label{boundphi1}\phi(x)\le C_2\frac{\sqrt{1-t^2}}{m},\quad x\in \tilde \Omega.
 \ee 
  Since the functions appearing in the sum for $g$ are all smaller than $\phi$ from (iii) of Lemma \ref{L:decomega}
  we conclude that
\be 
 |g(x)|\le   C_3\frac{\sqrt{1-t^2}}{m},\quad x\in \tilde \Omega.
 \ee 
This proves (i) and concludes the proof of the lemma.
\hfil $\Box$
\vskip .1in
\noindent{\bf  Proof of  Theorem \ref{T:approxphi}}  According to Remark \ref{R:proof}, we only need to consider  the case $\phi=\phi(\cdot;\xi,t)$ when $1/2\le  t<t_{2m-L}$.
We return to our representation \eref{yes}. As already mentioned $\phi(x)=g(x)$ outside
the set $\tilde \Omega$.  From Lemma \ref{L:error} it follows that
\be  
\label{wehave3}
\|\phi-g\|_{L_2(\Omega)}\le    C|\tilde \Omega|_d^{1/2}\frac{\sqrt{1-t^2}}{m}\le Cw(\phi)m^{-3/2}.
\ee 
Since $m^d\ge c_dn$, we have completed the proof of Theorem \ref{T:approxphi}.
\hfill $\Box$

\subsection{Proof of Theorem \ref{T:B^d}}
We can now prove Theorem \ref{T:B^d}
in the same way we proved the case $d=2$.    Let  $f$ be any fixed function from $\cV_w(\Omega)$. According to the definition of $\cV_w(\Omega)$, for $N$ sufficiently large, there is an $S\in\Sigma_N$ with $S=\sum_{j=1}^Na_j\phi_j$  such that
   \be
   \|f-S\|_{L_2(\Omega)}\le M n^{-\frac{1}{2}-\frac{3}{2d}}\quad {\rm and} \quad \sum_{j=1}^N w(\phi_j) |a_j|\le \|f\|_{\cV_w}=:M.
   \ee
   For each $j$, let 
    $g_j\in X_n$ approximate the function $\phi_j$ appearing in the representation of $S$ according to   Theorem \ref {T:approxphi}.
    That is, we have
    \be
    \label{wehave1}
    \|\phi_j-g_j\|_{L_2(\Omega)}\le C w(\phi_j) n^{-\frac{3}{2d}},
    \ee
    with $C$ depending only on $d$.  The function $g:=\sum_{j=1}^N a_jg_j$ is in $X_n$ and hence in $\Sigma_n$.  We write
    \be
    \label{writedecomp1}
    f= f-S+h+g,\quad h:=S-g=\sum_{j=1}^N a_j[\phi_j-g_j].
    \ee
      Therefore,
      \be
      \label{T111}
      E_{3n}(f)\le M n^{-\frac{1}{2}-\frac{3}{2d}} +E_{2n}(h).
      \ee 
      
      To bound $E_{2n}(h)$, we  consider the dictionary $\cD'=\{\psi_j\}_{j=1}^N$ with $\psi_j:= w(\phi_j)^{-1}(\phi_j-g_j)$.    According to Theorem \ref{T:approxphi}  each $\psi_j$ has $L_2(\Omega)$ norm at most $Cn^{-\frac{3}{3d}}$ and $h=\sum_{j=1}^Nc_j\psi_j$ with  $\sum_{j=1}^N|c_j|\le M$.  It follows from Maurey's Theorem (see \eref{Mtheorem}) that $h$ can be approximated by a sum $T$ of $n$ terms from the dictionary $\cD'$ with error
      \be
      \label{T121}
      \|h-T\|_{L_2(\Omega)} \le CMn^{-\frac{3}{2d}}n^{-1/2}=CMn^{\frac{1}{2}-\frac{3}{2d}}.
      \ee
         The function $T $ is a sum of at most $2n$ terms from the original dictionary $\cD$.  Hence,
      \be
      \label{T131}
      E_{2n}(h) \le CM n^{-\frac{1}{2}-\frac{3}{2d}}.
      \ee
    If we place this inequality back into \eref{T111}, we obtain 
   \be
      \label{T141}
      E_{3n}(f)\le  CM n^{-\frac{1}{2}-\frac{3}{2d}},
      \ee
  and the theorem follows.\hfill $\Box$

\begin{remark}
    \label{R:Qd}
If we consider $\Omega=Q^d$ in place of $B^d$ then for the weight $w(\phi(\cdot; \xi,t))= |L_\phi|^{1/2}$ where $|L_\phi|$ is the $(d-1)$-dimensional measure of the intersection $Q^d\cap L_\phi$, we obtain
\be 
\label{RQd}
E_n(f)_{L_2(Q^d)}\le Cn^{-\frac{1}{2}-\frac{3}{2d}} \|f\|_{\cV_w},\quad f\in\cV_w.
\ee 
We do not give the proof which follows along the lines of the case $d=2$ given in \S \ref{SS:Q2}.
\end{remark}

\input{conclusion}

     \section{Appendix}
     \label{S:appendix}
    In this appendix, we prove Lemma \ref{L:decomega}.  We let $\xi $ be arbitrary but fixed throughout this section.  We begin by recalling some well known results on the representation of points $x$ in a cube $R\subset \R^d$ in terms of the vertices of $R$.
     Given any cube $R\subset \R^d$, we let $V(R)$ denote its set of vertices.
 Let us first consider the case $R=U$ where  $U:=U^d:=[0,1]^d$.     We denote the vertices in $V(U)$ by $e$.
 So $e$ is a vector with $d$ components $e=(e_1,\dots,e_d)$ with each $e_j\in\{0,1\}$.  There are $2^d$ such $e$. 
 
 Let
 \be 
 \label{erep}
 \ell_{0}(s):= (1-s)\quad \ell_1(s)=s,\quad s\in \R.
 \ee 
 For each $e\in V$, we define
 \be 
 \label{defell}
 \ell_e(x):=\prod_{j=1}^d \ell_{e_j}(x_j),\quad  x=(x_1,\dots,x_d)\in U.
 \ee
 Then,  $\ell_e(e')=0$, $e'\neq e$ and $\ell_e(e)=1$.  Any $x\in U$ is represented as
 \be 
 \label{erep1}
 x=\sum_{e\in V}\ell_e(x) e.
 \ee
 This is a convex representation in that the coefficients $\ell_e(x)\in[0,1]$, $e\in V(U)$,  and they sum to one.

 Now consider an  arbitrary  cube $R\subset \R^d$.   We can write $R=v+\alpha [0,1]^d=v+\alpha U$ with $\alpha>0$.  This cube has vertices $v+ \alpha e$, $e\in V$.   Any point  $x=v+\alpha y$, $y\in U$,  from  this cube,
 has the representation  
 \be 
 \label{erep2}
 x=v+\alpha \sum_{e\in V}\ell_e(y) e=\sum_{e\in V} \ell_e(y)[v+\alpha  e],
 \ee
 because $\sum_{e\in V}\ell_e(y)=1$.
 Again this is the representation of $x$ as a convex combination of the vertices  $V(R)$ of $R$.
 Let us note that here we are taking $v$ as the smallest vertex of $R$.  We can derive a similar decomposition by starting with any other vertex of $R$.
 
 We use the above to find a variety of representations of any $x$ on the boundary of  the cube $Q^d:=[-1,1]^d$. Later, we shall apply these representations to $x=\bar\xi$ and subsequently to $\xi\in S^{d-1}$.   Let $x$ be in the face $F$ of $Q^d$. We assume that the   $d-1$ dimensional face $F$ of $Q^d$ corresponds to $x_1=1$.  
 We will derive representations for points $x\in F$.  Similar representations hold for any of the other faces of $Q^d$. 
 
 Any $x\in F$ takes the form $x=(1,\tilde x)$ with $\tilde x\in [-1,1]^{d-1}$.   
 Suppose now that $R$ is any $d-1$ dimensional cube on $F$, i.e., $R$ consist of points $(1,\tilde x)$ where $\tilde x$ is in a $d-1$ dimensional cube $\tilde R$.
      From the above, we can write $\tilde x=\tilde v +\alpha y$, $y\in U^{d-1}$, where $\tilde v$ is the smallest vertex of $R$.  Therefore, we have  
 \be 
 \label{xrep}
 \tilde x= \tilde v +\alpha y=\tilde v+\alpha \sum_{e\in V(U^{d-1})}\ell_e(y) e=\sum_{e\in V(U^{d-1})} \ell_e(y)[\tilde v+\alpha  e] =\sum_{\nu\in V(\tilde R)} \gamma_\nu \nu,
  \ee
  where the $\nu$ are the vertices of $\tilde R$ and
  \be 
  \label{gammas}
  \gamma_\nu =\ell_e(y),\quad {\rm when} \ \nu= \tilde v+ \alpha e.
  \ee
   This is a representation of $\tilde x$ as a convex combination of the vertices $V(\tilde R)$.

  We  turn now to representations of  $\xi\in S^{d-1}$. We write $\xi=\frac{\bar \xi }{\|\bar\xi\|}$ where $\bar\xi$ lies on the boundary of $Q^d=[-1,1]^d$. 
  We assume $\bar\xi $ lies on the face $F$ corresponding to first coordinate equal to one.  All other cases are handled similarly.
 We write $\bar \xi=(1,\tilde x)$ with $\tilde x\in [-1,1]^{d-1}$ and use the representations of $\tilde x$ given above.
 Recall the discrete set of points $F_k$, with $m=2^{k}$. If $k'<k$ then $F_{k'}\subset F_k$. We fix such a $k'$ to be chosen in a moment. 
 
 We let $A\ge 1$ be an integer whose value will be chosen below. We place ourselves in the following
 situation where $\tilde x \in \tilde v+\delta U^{d-1}=:\tilde R\subset \tilde R':=\tilde v+\delta' U^{d-1}$ where $\tilde v\in F_k$, $\delta=2^{-k}=1/m$  and
 $\delta'=2^{-k'}=A\delta$ with $A=2^{k-k'}$.    The assumption that  $\tilde x\in \tilde R$ for which there is such a $\tilde R$ and $\tilde R'$ is a restriction on the position of $\tilde x$ in $[-1,1]^{d-1}$.  When this is not the case, the argument below needs to be adjusted by changing the choice of the initial vertex and the direction for the representation.  Since the adjustment is purely notational, we leave it to the reader.

     We will give two representations of  $\tilde x$, respectively $\bar \xi$; the one in terms of the vertices of $\tilde R$ and the second in terms of the vertices of $\tilde R'$.
   For the first representation, we use \eref{xrep} with $\alpha=\delta$ to write
 \be 
 \label{repomegabar}
 \bar \xi= \sum_{\nu\in V(\tilde R)} \gamma_\nu (1,\nu)=\sum_{\nu\in V(\tilde R)} \gamma_\nu  \sqrt{(1+\|\nu\|^2)} \xi_\nu,\quad \xi_\nu= \frac{(1,\nu)}{\sqrt{1+\|\nu\|^2}},
 \ee
 with the coefficients $\gamma_\nu$  given by \eref{gammas}.  Notice that the $\xi_\nu$ are all in $W_k$. This gives the representation  
 \be 
 \label{repomega1}
  \xi=  \sum_{\nu\in V(\tilde R)} a_\nu \xi_\nu,\quad a_\nu:= \frac{\gamma_\nu  \sqrt{(1+\|\nu\|^2)} }{\|\bar\xi\|}.
 \ee

 We obtain  a second representation as follows. We again write $\tilde x=\tilde v+\delta y$ with $y\in U^{d-1}$.  Then,
 \be 
 \label{xrep1}
 \tilde x= \tilde v +\delta \sum_{e\in V(U^{d-1})}\ell_e(y) e =\tilde v + \sum_{e\in V(U^{d-1})}\frac{\ell_e(y)}{A}  A \delta e=\left(1-\frac{1}{A}\right)\tilde v+\sum_{e\in V(U^{d-1})} \frac{\ell_e(y)}{A}[\tilde v+A\delta e].
 \ee
  This gives the representation
  \be 
 \label{xrep2}
 \tilde x=  \sum_{\nu\in V(\tilde R')} \gamma_\nu ' \nu,
 \ee
where
\be 
  \label{gammas1}
  \gamma'_\nu = \frac{\ell_e(y)}{A},\quad {\rm when} \ \nu= \tilde v+ A\delta e \ {\rm with} \ e\neq 0\ {\rm and} \ \gamma'_{0}=1-\frac{1}{A} +\frac{\ell_0(y)}{A}.
  \ee
Notice that this representation of $\tilde x$ is again a convex combination of the vertices of $\tilde R'$.  It follows that

 \be 
 \label{repomega2}
  \xi=  \sum_{\nu\in V(\tilde R')} a_\nu' \xi'_\nu,\quad a'_\nu:= \frac{\gamma'_\nu  \sqrt{(1+\|\nu\|^2)} }{\|\bar\xi\|}.\quad \xi'_\nu= \frac{(1,\nu)}{\sqrt{1+\|\nu\|^2}}
 \ee
 
 We now want to estimate the  sums 
 \be
 \label{sums} S:=\sum_{\nu\in V(\tilde R)} a_\nu,\quad S'= \sum_{\nu\in V(\tilde R')} a'_\nu,
 \ee

 \begin{lemma} 
\label{L:compare}  There is an $m^*=m^*(d)$, depending only on $d$,  such that whenever $m\ge m^*$ and $A$ is sufficiently large (depending only on $d$), the following hold.
Whenever $\xi$ is not a vertex in $W_k$, i.e.,   $\bar\xi=(1,\tilde x)$ where $\tilde x=\tilde v+y$ where $y\neq 0$,  we have 
\be 
\label{compare} 
S=1+\epsilon\quad {\rm and} \quad S'= 1+\epsilon ',\quad {\rm where}\quad 0<2|\epsilon|<  |\epsilon'|\le  \sigma \delta^2,
\ee 
 and
    \be 
    \label{sigma}
    \sigma:=\sigma(y) =\sum_{e\neq 0} \ell_e(y)>0,
    \ee
    where the strict  inequality holds because $y\neq 0$.
\end{lemma}
\vskip .1in
\noindent

{\bf Proof:}  Let $B^2:=1+ \|\tilde v\|^2$ and recall that $\delta:=1/m$.  Observe that when $\nu=\tilde v+a\delta e$, $e\in V(U^{d-1})$, we have
  \be 
  \label{esta0}
  1+  \| \nu\|^2=B^2+ 2a\delta  \langle \tilde v,e\rangle +a^2\delta^2\|e\|^2=B^2+s_\nu(a), 
    \ee
    where 
    \be
    \label{snu}
    s_\nu(a):= 2a\delta \langle \tilde v,e\rangle +a^2\delta^2\|e\|^2.
    \ee
   We are interested in the cases, $a=1,A$.  Notice that $s_\nu(a)=0$ when $e=0$, i.e., $\nu=\tilde v$, and also $|s_\nu(a)|\le 1/2$ for these two values of $a$ provided $m^*$ is large enough.  These facts will be used without further mention in what follows.
    
We will use the Taylor expansion
 of the function $F(s):= \sqrt{B^2+s}$.  We have
 \be
 \label{taylor}
 F(s)=  B+ \frac{1}{2}B^{-1}s- \frac{1}{4} B^{-3}s^2+ O(s^3),\quad |s|<1.
 \ee
 This gives that
 \be 
 \label{observe}
 \sqrt{1+\|\nu\|^2}=F(s_\nu(a)) =B+ \frac{1}{2}B^{-1}s_\nu(a)- \frac{1}{4} B^{-3}s_\nu(a) ^2+ O(s_\nu(a)^3)
\ee 
From the above  observations, we can write
    \begin{eqnarray} 
    \label{S2}
    \|\bar\xi\|S'&=&\sum_{\nu\in V(\tilde R')} \gamma'_\nu F(s_\nu(A))\nonumber\\
    &=&(1-1/A)B+\sum_{e\in V(U^{d-1} )}\frac{\ell_e(y)}{A}\left(B+ \frac{1}{2}B^{-1}s_\nu(A)- \frac{1}{4} B^{-3}s_\nu(A)^2+ O(s_\nu(A)^3)\right)\nonumber\\
     &=& B+ \frac{B^{-1}}{2}\sum_{e\in V(U^{d-1})}\frac{\ell_e(y)s_\nu(A)}{A} - \frac{B^{-3}}{4}\sum_{e\in V(U^{d-1} )}\frac{\ell_e(y)s_\nu(A)^2}{A}+ O(A^2\sigma \delta^3).
     \end{eqnarray}

     Let us analyze the first sum $\Sigma_1$ in \eref{S2}.  Using the definition of the $s_\nu$, we see that this sum equals
     \be 
     \label{firstsum} 
    \Sigma_1= C_1\delta+C_2A\delta^2,\quad {\rm where} \quad C_1=B^{-1}\sum_{e\neq 0} \ell_e(y)\langle \tilde v,e\rangle \ {\rm and}  \ C_2=\frac{B^{-1}}{2}\sum_{e\neq 0}\ell_e(y)\|e\|^2 .
     \ee 
     A similar analysis of the second sum $\Sigma_2$ gives
 \be 
     \label{secondsum} 
    \Sigma_2= C_3A\delta^2+C_4A^2\delta^3+C_5A^3\delta^4,\quad {\rm where}  \quad C_3=B^{-3}\sum_{e\neq 0}\ell_e(y)\langle \tilde v,e\rangle ^2, \quad {\rm and} \ |C_4|,|C_5|\le C_0\sigma,
     \ee
     where $C_0$ depends at most on $d$.
   In total, this gives
   \be 
   \label{totalS}
   \|\bar\xi\|S'=   B+ C_1\delta + \tilde C \sigma A\delta^2 +O(\sigma A^2\delta^3),
   \ee 
   where 
   \be 
   \label{tildeC} 
   \sigma\tilde C= C_2 +C_3,
   \ee
   and where the constants in the "$O$" term depend only on $d$.
   It is important to notice that 
   \be 
   \label{noticeC}
   \tilde C\ge \sigma^{-1} C_2 \ge 1/4.
   \ee 
   Replacing $A$ by one, we get
   \be 
   \label{totalS1}
   \|\bar\xi\|S=   B+ C_1\delta +\tilde C \sigma \delta^2 +O(\sigma  \delta^3).
   \ee 
Notice that these constants   are the same as those in \eref{totalS} and again the constants in the "$O$" term depend only on $d$.

Next, we want to compute $\|\bar\xi\|$ and compare this number with $B+C_1\delta$.  We have $\bar\xi=(1,\tilde x)$
where 
$$\tilde x= \tilde v+\delta y= \tilde v +\delta\sum_{e\in V(U^{d-1})} \ell_e(y)e.$$
Therefore,
\be 
\label{normtildeomega}
\|\tilde \xi\|^2= 1+\|\tilde v\|^2 +2\delta \sum_{e\in V(U^{d-1})}\ell_e(y)\langle \tilde v,e\rangle +\delta^2\|y\|^2 = B^2+s.
\ee

    If we  now use \eref{taylor}, we obtain
$$\|\bar\xi\|= F(s)=B+\frac{B^{-1}}{2}s-\frac{B^{-3}}{4}s^2 + O(s^3)= B+C_1\delta+ \tilde C'\sigma \delta^2 + O( \sigma \delta^3).
$$
where 
\be 
\label{next C}
\sigma \tilde C'= \frac{B^{-1}}{2}\delta^2\|y\|^2+ B^{-3}\left[\sum_{e\neq 0}\ell_e(y) \langle \tilde v,e\rangle\right]^2\delta^2.
\ee 
Here, we have also used the fact that   $\|y\|\le \sigma$.    If we use this expression for $\|\bar\xi\|$ in \eref{totalS1}, we obtain
\be 
\label{Cstar} 
S= 1+C^* \sigma \delta^2+ O(\sigma\delta^3)=:1+\e, \quad C^*=\tilde C-\tilde C'.
\ee 
Similarly
\be 
\label{Cstar1} 
S'= 1+C^{**} \sigma \delta^2+ O(\sigma A^2\delta^3)=:1+\e', \quad  C^{**}=\tilde C  A^2-\tilde C' .
\ee 
If we choose $m$ sufficiently large ($m\ge m^*$ with $m^*$ depending only on $d$ and $A$ as a sufficiently large integer depending only on $d$ we will have $0<2\e<\e'$ (see \eref{noticeC}). This completes the proof of the Lemma.
    \hfill $\Box$

    Note that the constant $A$ of this lemma serves to define $A$ for this paper and then $L=(A+1)^2$ is defined as in Lemma \ref{L:ti+}.

\vskip .1in
\noindent
{\bf Proof of Lemma \ref{L:decomega}:}  
\vskip .1in
\noindent
{\bf Case $\xi\in W_k$:}  Let $\xi=\xi_i\in W_k$.  
Given $t\in [1/2,t_{2m-L}]$,  we have $t_i^+=t_j$ and we take
$\tilde t_i:=t_{j+1}$.  We define $\alpha$ by the requirement
\be 
\label{defalpha}
\alpha t_i^++(1-\alpha)\tilde t_i =t, \quad {\rm i.e.} \quad \alpha =\frac{t-\tilde t_i}{t_i^+-\tilde t_i}.
\ee
Then, $\xi = \alpha \xi +(1-\alpha) \xi$, which is  the decomposition for $\xi$ required in Lemma \ref{L:decomega}.
Indeed, $|\alpha|\le C$ with $C$ depending only on $d$ because  of Lemma \ref{L:ti+}  and \eref{factspacing}.

\vskip .1in
\noindent
{\bf Case $\xi$ is not in $W_k$:}     
 We will use the constructions given above. We take $A$ to be an integer as given in Lemma \ref{L:compare}. We have given two ways of representing $\xi$ as given in \eref{repomega1} and \eref{repomega2}. The $\xi_\nu$ and $\xi_\nu'$ appearing in these representations are all from $W_k$.   We take $W_k(\xi)$ as the collection of all these points.  
Property (ii) of Lemma \ref{L:decomega} is satisfied  since $\|\xi-\xi_i\|\le A/m$ for each $i$.
We define $\alpha$ by the requirement
 \be 
\label{defbeta}
\alpha \epsilon+(1-\alpha)\epsilon '=0, \quad {\rm i.e.} \quad \alpha =\frac{\epsilon ' }{\epsilon'-\epsilon }.
\ee
It follows that 
\be 
\label{lastomega}
\xi = \alpha  \sum_{\nu\in V(R)} a_\nu \xi_\nu+(1-\alpha) \sum_{\nu\in V(R')} a'_\nu \xi'_\nu,
= \sum_{j=1}^M b_j\xi_j,\quad \sum_{j=1}^M b_j=1,
\ee
where all of the $\xi_j$ are in $W(\xi)$.  The key here is that the coefficients in this representation sum to one.

 Now, given $t\in[1/2,t_{m-L}]$, we define
 \be 
 \label{deft+tilde}
 t^+:=\max \{t_i^+: \ \xi_i\in W_k(\xi)\}=t_j,\quad \tilde t: =t_{j+1}.
 \ee 
 Similar to the above, we define $\beta$ by requiring that 
 \be 
\label{defalpha1}
\beta t^++(1-\beta)\tilde t =t, \quad {\rm i.e.} \quad \beta =\frac{t-t^+}{t^+-\tilde t}.
\ee
It follows that
\be 
\label{lastomega1}
\xi\cdot x-t =  
 \sum_{j=1}^M \beta b_j(\xi_j\cdot x-t^+)+\sum_{j=1}^M (1-\beta) b_j(\xi_j\cdot x-\tilde t). 
\ee
 This is the decomposition promised in Lemma \ref{L:decomega} and thereby completes the proof of the lemma.
 \hfill $\Box$

\bibliographystyle{plain}
 \bibliography{refs}

\vskip .1in
\noindent 
Ronald DeVore, Department of Mathematics, Texas A\&M University,
College Station, TX 77843
\vskip .1in
\noindent 
Robert D. Nowak, Department of Electrical and Computer Engineering, University of Wisconsin--Madison,  Madison, WI 53706
\vskip .1in
\noindent
Rahul Parhi, Department of Electrical and Computer Engineering, University of California, San Diego, La Jolla, CA 92093
\vskip .1in
\noindent 
 Jonathan W. Siegel, Department of Mathematics, Texas A\&M University,
College Station, TX 77843
 
 \end{document}

%% file: conclusion.tex
\section{Concluding Remarks}
\label{S:concluding}

The main theme of this  paper was to  introduce, for a bounded domain $\Omega\subset \R^d$,  new model
 classes $\cV_w := \cV_w(\cD(\Omega)$), called {\it weighted variation spaces}, and to prove bounds on how well functions in these classes can be
 approximated by a linear combination of $n$ terms of the ReLU dictionary $\cD(\Omega)$. That is, we provided bounds on the error $E_n(f)_{L_2(\Omega)}$
 in approximating $f\in \cV_w(\Omega)$  in the $L_2(\Omega)$ norm by the elements of $\Sigma_n:=\Sigma_n(\cD(\Omega))$ where $\Sigma_n$ is the nonlinear manifold
 of functions $g$ that are a linear combination of $n$ elements of the ReLU dictionary. We showed that for certain choices of the weight $w$ (dependent on $\Omega$)
 the functions in these new model classes have the same approximation rate as those in the classical variation spaces
 $\cV(\Omega)$.  Since $\cV_w$ is strictly  larger than the classical variation spaces $\cV:=\cV(\Omega)$, this gives stronger results
 on $n$-term ReLU approximation than those in the literature.  Thus, these new model classes $\cV_w$ are important in trying to understand which functions are well approximated by $\Sigma_n$.

A natural follow-up question would be to then consider the problem of learning from data generated from from the samples of a function from $\cV_w$, in both the noiseless and noisy settings. In the literature, the former is referred to as \emph{optimal recovery} and the latter is referred to as \emph{minimax estimation}. For the classical variation spaces, the minimax estimation rates have been determined~\cite{PNMinimax}. On the other hand, the optimal recovery rates are currently unknown. Once the data sites are fixed, it is well-known that the procedure for optimal recovery takes the form of solving a regularized least-squares problem over the model class~\cite{binev2024optimal}. Theorem~\ref{representer} below motivates a numerical method (posed as a neural network training problem) to investigate the problem of optimal recovery (as well as for minimax estimation for $\cV_w$).

Assume that $d\ge 2$ and $\Omega = B^d$ in the sequel. Then, $\bar Z(\Omega)=S^{d-1}\times [-1,1]$. Let $w$ be any admissible weight function in the sense of \eqref{wa}  
and let $\cD_w$ be the weighted dictionary defined in \eref{newdictionary}.  We use the results and notation of \S \ref{S:variation}  in going forward.  In particular, the functions in $\cV_w:=\cV_w(\Omega)$ all take the form (see \eref{weightrep})
\be 
\label{form}
\tilde f_\mu:=\int_{\bar Z(\Omega)} \tilde \phi(\cdot;\xi,t)\, d\mu\quad {\rm and} \quad \|\tilde f_\mu\|_{\cV_w(\Omega)}= \|\mu\|_{\cM}.
\ee

Consider the following data-fitting problem.  Suppose that $x_i$, $i=1,\dots,m$, are points from the interior of $\Omega$
and $y_i$, $i=1,\dots, m$, are real numbers.  The data-fitting problem
\begin{equation}
    \inf_{f \in \cV_w(\Omega)} \sum_{i=1}^m |y_i - f(x_i)|^2 + \lambda \norm{f}_{\cV_w},
    \label{eq:opt-Vw}
\end{equation}
with $\lambda > 0$ is equivalent to the data-fitting problem
\begin{equation}
    \inf_{\mu \in \cM(\bar Z(\Omega))} \sum_{i=1}^m \left|y_i -  f_\mu(x_i) \right|^2 + \lambda \norm{\mu}_{\cM},
    \label{eq:opt-measures}
\end{equation}
in the sense that their infimal values are the same and if $\mu^\star$ is a minimizer of \eqref{eq:opt-measures}, then
$f_{\mu^*}$
is a minimizer of \eqref{eq:opt-Vw}. Note that the minimization problem \eref{eq:opt-Vw} does not depend on the ambient space $L_2(\Omega)$ in which we measure error of performance for the approximation problem. An important property of the weighted variation spaces is that solutions to data-fitting problems over this model class admit finite-parameter representations as neural networks. This is summarized in the next theorem.

\begin{theorem} \label{representer}
    Suppose that $w$ is an admissible weight function and  the  $\{x_i\}_{i=1}^m$ lie in the interior of $\Omega$. Then, there exists a solution $f^*$ to \eqref{eq:opt-Vw} that takes the form of a shallow ReLU network
    \begin{equation}
        f^\star(x) = \sum_{j=1}^n a_j \phi(x; \xi_j, t_j) = \sum_{j=1}^n a_j (\xi_j \cdot x - t_j)_+,
    \end{equation}
    where the number of atoms satisfies $n \leq m$, $\{a_j\}_{j=1}^n \subset \R \setminus \{0\}$, and $\{(\xi_j, t_j)\}_{j=1}^n \subset \bar Z(\Omega)$ are data-dependent and not known \emph{a priori}. Furthermore, the regularization cost is $\norm{f^\star}_{\cV_w} = \sum_{j=1}^n w(\xi_j, t_j) \abs{a_j}$.
\end{theorem}
\vskip .1in
\noindent
{\bf Proof:}
Let $C(\bar Z(\Omega))$ denote the space of real  valued functions   on $\bar Z(\Omega)$. This is a Banach space when equipped with the  $L_\infty$-norm. By the Riesz--Markov--Kakutani representation theorem~\cite[Chapter~7]{FollandRA}, the dual of $C(\bar Z(\Omega))$ can be identified with the space of signed Radon measures $\cM:=\cM(\bar Z(\Omega))$. It is well-known that the extreme points of the unit ball
\be
    \{\mu \in \cM : \norm{\mu}_{\cM} \leq 1\}
\ee
are the Dirac measures $\pm \delta_{(\xi, t)}$, $(\xi, t) \in \bar Z(\Omega)$ (see, e.g.,~\cite[Proposition~4.1]{BC}).

Next, for $i = 1, \ldots, m$, we introduce the functions
\be 
\label{defhi}
h_i(\xi, t) := \tilde \phi(x_i;\xi,t)=\frac{\phi(x_i; \xi, t)}{w(\xi, t)},\quad (\xi,t)\in \bar Z(\Omega). 
\ee 
We can rewrite \eqref{eq:opt-measures} as
\be
    \inf_{\mu \in \cM } \sum_{i=1}^m |y_i - \langle \mu, h_i \rangle |^2 + \lambda \norm{\mu}_{\cM},
    \label{eq:opt-measures2}
\ee
where $\langle \cdot, \cdot \rangle$ denotes the duality pairing between $C(\bar Z(\Omega))$ and $\cM$. Since the functions $h_i $, $i=1,\dots,m$, are in $ C(\bar Z(\Omega))$,   the mappings $\mu \mapsto \langle \mu, h_i \rangle$ are weak$^*$ continuous~\cite[Theorem~IV.20, p. 114]{RSBook}. This shows that the hypothesis of the abstract representer theorem~\cite{BCDDDW,BC,U} are satisfied.  That theorem shows  that there exists a solution to \eqref{eq:opt-measures2} that takes the form of a linear combination of the extreme points of the unit regularization ball. Thus, there exists a solution that takes the form
\be
    \mu^\star = \sum_{j=1}^n c_j \delta_{(\xi_j, t_j)},
\ee
where the number of atoms satisfies $n \leq m$, $\{c_j\}_{j=1}^n \subset \R \setminus \{0\}$, and $\{(\xi_j, t_j)\}_{j=1}^n \subset \bar Z(\Omega)$ are distinct, data dependent, and not known \emph{a priori}. Clearly $\norm{\mu^\star}_{\cM} = \sum_{j=1}^n \abs{c_j}$.

From the equivalence between \eqref{eq:opt-Vw} and \eqref{eq:opt-measures2}, we see that the function 
\begin{equation}
\label{solution}
    f_{\mu^\star} =  \int_{\bar Z(\Omega)}  \tilde \phi(\cdot; \xi, t)  d\mu^\star(\xi, t) = \sum_{j=1}^n \frac{c_j}{w(\xi_j, t_j)} \phi(x; \xi_j, t_j)
\end{equation}
is a minimizer of \eqref{eq:opt-Vw}. The theorem follows by the substitution $a_j := c_j / w(\xi_j, t_j)$.
\hfill $\Box$

We have not indicated the fact that the solution \eref{solution} to the data-fitting problem depends on $\lambda$.  If we let
$\lambda$ tend to zero then the solutions converge to a minimum-norm interpolant $f^{\#}$ of the data
\be 
\label{minnorm}
f^{\#} \in \argmin \{\|f\|_{\cV_w}: f(x_i)=y_i,\ i=1,\dots,m\}.
\ee 
In which case, there always exists an $f^{\#}$ that has a representation 
\be 
\label{minnormrep} 
f^{\#} =   \sum_{j=1}^n  a_j^{\#}  \phi(x; \xi_j^{\#}, t_j^{\#}),
\ee 
with $n\le m$.
 
The theorem statement also holds when the first term in the objective in \eqref{eq:opt-Vw} is replaced by any loss function $\cL(\cdot, \cdot)$ which is lower semi-continuous (see~\cite[Proof~of~Theorem~3.2]{PNDeep}). In neural network parlance, the $\xi_j$ are referred to as the \emph{input weights}, the $a_j$ are referred to as the \emph{output weights} and the $t_j$ are referred to as the \emph{biases}. Observe that the norm of a single neuron $\phi(\cdot; \xi, t)$, where $\xi \in \R^d$ and $t \in \R$, takes the form
\begin{equation}
    \norm{\phi(\cdot; \xi, t)}_{\cV_w} = \norm{\xi} w\left(\frac{\xi}{\norm{\xi}}, \frac{t}{\norm{\xi}}\right),
\end{equation}
where we took advantage of the fact that the ReLU is positively homogeneous of degree $1$. In this form,  the input weights are not restricted to be unit norm. Theorem~\ref{representer} then implies that a solution to the variational problem in \eqref{eq:opt-Vw} can be found by training a sufficiently wide (fixed width $n \geq m$) neural network to a global minimizer with an appropriate regularization term. This follows, in particular, by finding a solution to the neural network training problem
\begin{equation}
    \min_{\theta} \sum_{i=1}^m |y_i - f_\theta(x_i)|^2 + \lambda \sum_{j=1}^n |a_j| \norm{\xi_j} w\left(\frac{\xi_j}{\norm{\xi_j}}, \frac{t_j}{\norm{\xi_j}}\right), \label{eq:nn-prob}
\end{equation}
where
\begin{equation}
    f_\theta(x) = \sum_{j=1}^n a_j \phi(x; \xi_j, t_j) = \sum_{j=1}^n a_j (\xi_j \cdot x - t_j)_+
\end{equation}
is a shallow ReLU neural network and $\theta = (a_j, \xi_j, t_j)_{j=1}^n$ denotes the neural network parameters and $n \geq m$.
When $\lambda$ is chosen to be sufficiently small, the estimator $f_{\widetilde{\theta}}$ achieves the optimal recovery rate for the model class $\cV_w$, where $\widetilde{\theta}$ is any minimizer of \eqref{eq:nn-prob}~\cite{binev2024optimal}. 

When $w$ is the weight specified in \eqref{defweightd} (which satisfies the hypotheses of Theorem~\ref{representer}), the resulting regularizer takes the form
\begin{equation}
    \sum_{j=1}^n |a_j| \norm{\xi_j} \left(1 - \frac{t_j}{\norm{\xi_j}}\right)^{\frac{1}{2} + \frac{d}{4}} \label{eq:new-reg}
\end{equation}
This is a new regularizer for training neural networks which directly penalizes the biases. If we assume the data sites $\{x_i\}_{i=1}^m$ are drawn i.i.d.\ uniformly on $B^d$, then this penalization reflects the volume of the subset of $B^d$ where the neuron is ``active''  (nonzero output). This suggests a new, data-adaptive regularization scheme in which the the penalty on a given neuron is proportional to the number of data in its support. This regularizer should be contrasted with the unweighted case in which the regularizer takes the form
\begin{equation}
    \sum_{j=1}^n |a_j| \norm{\xi_j},
\end{equation}
which is sometimes referred to as the \emph{path-norm}~\cite{NSS} of the neural network. Remarkably, path-norm regularization is equivalent to the common procedure of training a neural network with \emph{weight decay}~\cite{KH} which corresponds to a regularizer of the form
\begin{equation}
    \frac{1}{2}\sum_{j=1}^n |a_j|^2 + \norm{\xi_j}^2.
\end{equation}
We refer the reader to~\cite{PNSurvey} for more details about this equivalence. The new regularizer in \eqref{eq:new-reg} requires further study in both theory and practice.

\subsection{Open Problems}
The results presented in this paper open the door to several new research directions.
\begin{enumerate}
    \item We have shown that the classical variation space $\cV(\Omega)$ is not the approximation space $\cA^\alpha = \cA^\alpha(L_2(\Omega))$, $\alpha = \frac{1}{2} + \frac{3}{2d}$, since the (strictly larger) weighted variation space $\cV_w(\Omega)$ admits the same $n$-term approximation rate with shallow ReLU networks. Thus, the results of this paper bring us one step closer to characterizing the approximation space $\cA^\alpha$, $\alpha = \frac{1}{2} + \frac{3}{2d}$. Future work will be devoted to finding a characterization of this approximation space.

    \item The results of this paper only consider $L_2$-approximation. We conjecture that the same rates hold for weighted variation spaces for all $L_p$, $1 \leq p \leq \infty$, where now the admissibility condition on the weights will depend on $p$. That is to say, for each $1 \leq p \leq \infty$, there exists a weight function $w^*_p$ such that the the optimal rate  $n^{-\frac{1}{2}-\frac{3}{2d}}$ is achieved.

    \item 
    The determination of the optimal recovery rates and minimax estimation rates for $\cV_w$ is a natural follow-up research direction. Theorem~\ref{representer} and \eqref{eq:nn-prob} provide a numerical method (posed as a neural network training problem) whose solutions are known to achieve the optimal recovery rate. A characterization of this rate is has not been determined, even in the unweighted scenario.
     
    \item The weighted variation spaces motivates a new form of data-adaptive regularization for neural networks. Theoretical and experimental comparisons of this new form of regularization compared with more conventional regularization techniques is a direction of future work. Furthermore, extensions of this regularizer to deep neural networks is also a direction of future work.
\end{enumerate}